\documentclass[11pt]{article}
\usepackage[final]{acl}

\usepackage[T1]{fontenc}
\usepackage{times}
\usepackage{latexsym}
\usepackage[utf8]{inputenc}
\usepackage{inconsolata}
\usepackage{graphicx}
\usepackage{booktabs}
\PassOptionsToPackage{hyphens}{url}
\usepackage{url}
\usepackage{caption}
\usepackage{adjustbox}
\usepackage{siunitx}
\usepackage[table,dvipsnames]{xcolor}
\usepackage{pifont}
\usepackage{colortbl}
\usepackage{makecell}
\usepackage{amsmath}
\usepackage{amssymb}
\usepackage{arydshln}
\usepackage{array}
\usepackage[most]{tcolorbox}
\usepackage{enumitem}
\usepackage{multirow}

\usepackage{fancyhdr}
\newcommand{\aclfootershift}{-0.5pt}
\newcommand{\aclpagenumshift}{11.5pt}
\fancypagestyle{aclfirstpage}{%
  \fancyhf{}%
  \fancyfoot[C]{\raisebox{\aclfootershift}[0pt][0pt]{\parbox{\textwidth}{\centering
    {\normalsize\thepage\par}\vspace{3pt}
    {\footnotesize\setlength{\baselineskip}{11pt}%
     \textit{Proceedings of the 2026 Conference on Empirical Methods in Natural Language Processing}, pages 12612--12633\\
     October 24-29, 2026 \textcopyright 2026 Association for Computational Linguistics\par}}}}%
}
\fancypagestyle{aclotherpages}{%
  \fancyhf{}%
  \fancyfoot[C]{\raisebox{\aclpagenumshift}[0pt][0pt]{\normalsize\thepage}}%
}
\DeclareUnicodeCharacter{202F}{\,}
\newcommand{\cmark}{\textcolor{green!70!black}{\ding{51}}}

\newcommand{\dn}[1]{\textsubscript{\textcolor{red}{$\blacktriangledown$}#1}}
\newcommand{\up}[1]{\textsubscript{\textcolor{ForestGreen}{$\blacktriangle$}#1}}

\newcommand{\unknown}{\ifmmode\text{``Unknown''}\else``Unknown''\fi}

\newcommand{\blfootnote}[1]{{\renewcommand{\thefootnote}{}\footnotetext{#1}}}

\title{LLM Abstention Can Be a Prompt Artifact, \\in Addition to Genuine Uncertainty}

\author{
  Zipeng Ling$^{1,2,3}$\thanks{Equal Contribution},
  Shuliang Liu$^{1}$\footnotemark[1],
  Yuehao Tang$^{1}$,
  Junqi Yang$^{4}$,
  Shenghong Fu$^{5}$,\\
  \textbf{
  Seonil Son$^{6}$,
  Chen Huang$^{4}$,
  Kejia Huang$^{7}$,
  Yao Wan$^{4}$,
  Zhichao Hou$^{8}$,
  Xuming Hu$^{1}$}\thanks{Corresponding Author} \\
  $^{1}$The Hong Kong University of Science and Technology (Guangzhou)\\
  $^{2}$University of Alberta
  $^{3}$Alberta Machine Intelligence Institute\\
  $^{4}$Huazhong University of Science and Technology  $^{5}$The Hong Kong Polytechnic University\\
  $^{6}$RLWRLD  $^{7}$Nanyang Technological University $^{8}$University of Pennsylvania\\
    \texttt{\{zpling0816, xuminghu97\}@gmail.com}
}

\begin{document}
\setcounter{page}{12612}
\pagestyle{aclotherpages}
\maketitle
\thispagestyle{aclfirstpage}
\blfootnote{Code and datasets are available at the \href{https://github.com/ZpLing/EMNLP2026_Abstention-Inflation}{GitHub repository}}

\begin{abstract}
Large Language Models (LLMs) are increasingly trained to abstain from answering questions they are unsure about. However, this ability is often misapplied: in real-world applications, user prompts sometimes contain elements of uncertainty, which lead LLMs to abstain even on problems they are capable of solving. We argue that LLM abstention is not only an expression of genuine uncertainty; it can also be an artifact largely shaped by prompts. We name this phenomenon \emph{Abstention Inflation}. We add ``Unknown'' as an extra option for LLMs to choose from; experiments show serious accuracy drops on True/False Questions (TFQs). Replacing ``Unknown'' with an unrelated random word produces a similar effect. We argue that LLMs are trained to imitate the surface pattern of abstention, rather than to express genuine uncertainty. Based on ten experimental settings, we support four claims that form a progressive argument: \textbf{(C1)} \emph{Abstention Inflation} can be triggered by the presence of an extra option, not by genuine uncertainty; \textbf{(C2)} it makes the models deny they can answer, even when they can; \textbf{(C3)} it is a later-layer output override, as the reasoning traces and mid-layer representations preserve correct answers; \textbf{(C4)} it is not stochastic noise: it results from various factors, emerges through instruction tuning, is boosted by problems' higher difficulty, and can be mitigated at larger model sizes.
\end{abstract}

\section{Introduction}
\label{sec:intro}
LLMs are increasingly trained to abstain on questions they are unsure about. Since problems can sometimes be hard, offering overly confident answers may mislead users, especially in high-stakes domains such as medical treatment~\cite{machcha2026knowingabstainmedicalllms}, financial analysis~\cite{hou2026finsafetybenchevaluatingllmsafety}, and legal consultation~\cite{hu2025finetuninglargelanguagemodels}.


However, if LLM abstention comes not from genuine uncertainty but from a trigger (e.g., an ``Unknown'' option added to the prompt), it poses risks in several scenarios:
(1)~Question Answering~\cite{ren2023selfevaluationimprovesselectivegeneration}: users may express uncertainty along with questions, and such phrasing can affect LLM behavior. (2)~Benchmark Evaluation: adding ``Unknown'' as an option may introduce biases in LLMs' responses, yet in some fields it is important to include objectively unknown categories (e.g., factual question answering).

Existing research on LLM abstention treats outputting ``Unknown'' as a positive signal: if a model chooses to abstain, it knows to refuse proactively~\cite{ren2023selfevaluationimprovesselectivegeneration,zhang2024calibratingconfidencelargelanguage}; AbstentionBench~\cite{kirichenko2025abstentionbenchreasoningllmsfail} benchmarks how LLMs abstain for different reasons (e.g., underspecified context or question). However, we ask the reverse question: can LLM abstention also be a negative signal that harms accuracy, and can it be influenced by the input prompt? This question remains underexplored.

\begin{figure}[!t]
  \centering
  \includegraphics[
      width=\columnwidth,       
      keepaspectratio,
      trim=0 00pt 0 0,        
      clip
  ]{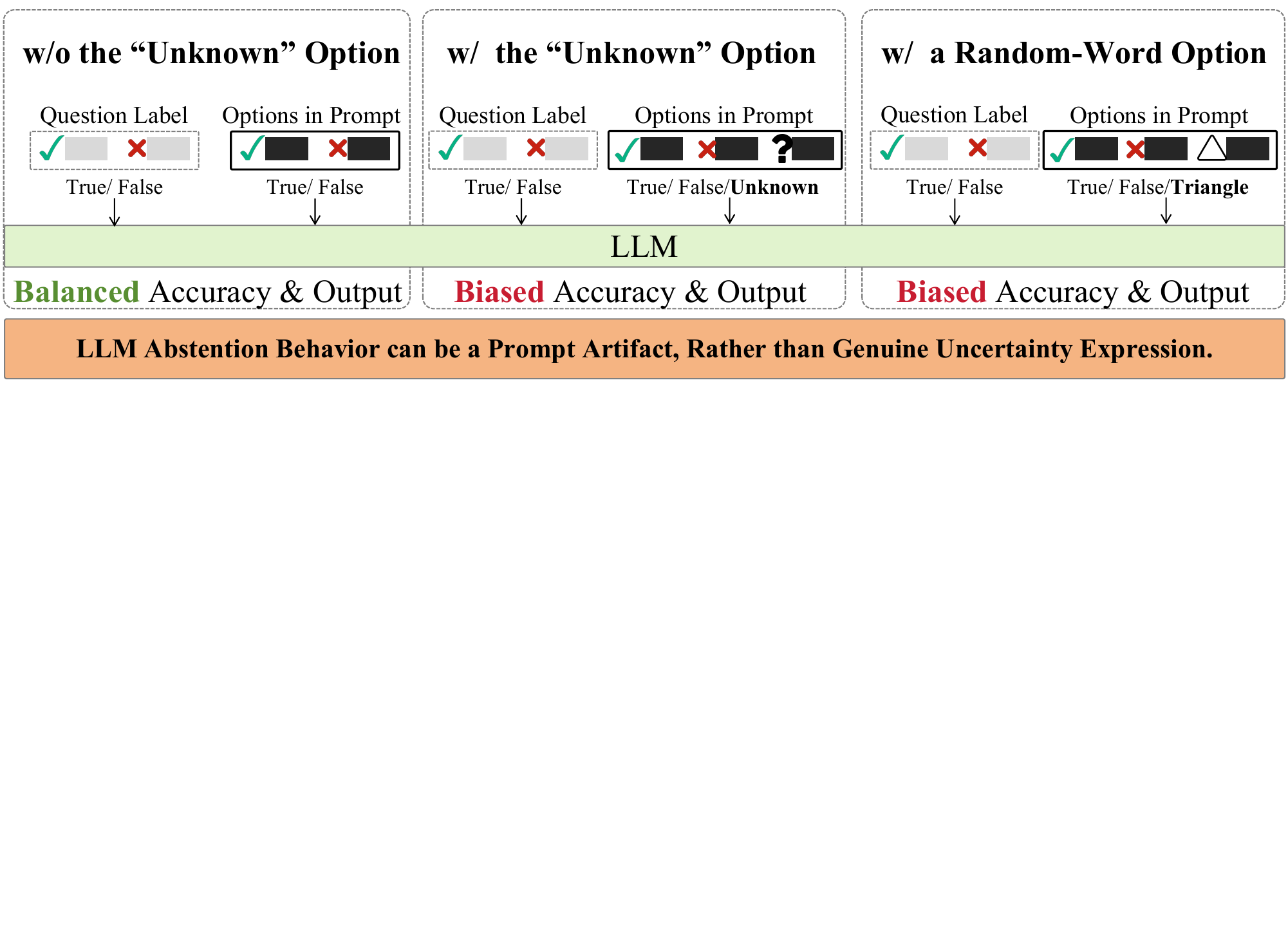}
  \vspace{-110pt}
\caption{Problem background: adding an ``Unknown'' option makes LLMs abstain heavily, and replacing it with a random word makes LLMs treat the word as a placeholder for ``Unknown'' and choose it. A detailed explanation and examples are in Section~\ref{sec:structure}.}

  \label{fig: IB_Formulation}
  \vspace{-10pt}
\end{figure}

 To study this, we add ``Unknown'' as an extra option to True/False Questions (TFQs) and Multiple-Choice Questions (MCQs): on TFQs, this causes LLMs to choose ``Unknown'' frequently, even on problems that can be correctly solved when this option is removed; on MCQs, it has little effect. Experiments show that when answering TFQs, LLMs are inclined to imitate the surface pattern of \emph{abstention}, in addition to genuinely expressing uncertainty. Based on this, we carry out systematic experiments analyzing the phenomenon.

We name this phenomenon \emph{Abstention Inflation}. Our contributions are:

\noindent$\triangleright$  We identify \emph{Abstention Inflation} as a failure mode of LLM abstention: on TFQs, adding an ``Unknown'' option causes models to abstain at scale, even on questions they can correctly answer. This comes from a prompt artifact, not from the LLM expressing genuine uncertainty, and is overlooked by existing abstention benchmarks.

\noindent$\triangleright$ Based on ten experiments, we establish four claims about \emph{Abstention Inflation}: (i)~it can be triggered by the presence of an extra option, not by genuine uncertainty; (ii)~it makes the models deny they can answer, even when they can; (iii)~it is a later-layer output override: the reasoning trace and mid-layer representations preserve the correct answer; and (iv)~it is not stochastic noise: it results from various factors, emerges through instruction tuning, is boosted by problems' higher difficulty, and can be mitigated at larger model sizes.

\noindent$\triangleright$ We argue that LLM abstention can be an invalid uncertainty signal, and offer advice on mitigating the problem. Our work helps in understanding LLM uncertainty expression.

\section{Related Work}
\label{sec:related}

\paragraph{LLM Abstention.}
Prior work on LLM abstention focuses on when models should refuse~\cite{wen2025knowlimitssurveyabstention,muhamed2025refusalbenchgenerativeevaluationselective,kirichenko2025abstentionbenchreasoningllmsfail} and whether such behavior tracks accuracy~\cite{zhang2024calibratingconfidencelargelanguage,huang2024uncertaintylanguagemodelsassessment,ren2023selfevaluationimprovesselectivegeneration}. Both lines of work assume that LLM abstention is faithful to the model's uncertainty. The most closely related work is AbstentionBench~\cite{kirichenko2025abstentionbenchreasoningllmsfail}, which reports a 24\% drop in abstention performance under fine-tuning. Our work studies the same behavior, but interprets it differently: we show that the same abstention behavior can also be negative, as an artifact triggered by the presence of an extra option on problems the model can solve.

\paragraph{Knowledge Boundary.}
\citet{Wen2024PerceptionKnowledgeBoundary} and \citet{li2025knowledgeboundarylargelanguage} study how models perceive their limits, and \citet{kadavath2022languagemodelsmostlyknow} show that LLMs can predict their own accuracy to some extent, based on the logits of true-label probabilities. We complement this at the empirical level: once an extra ``Unknown'' option is added, we find that the model's self-reported reason for abstaining can diverge from its true capability.


\begin{figure*}[t]
  \centering
  \includegraphics[
      width=\textwidth,
      keepaspectratio,
      trim=0 0 0 0,
      clip
  ]{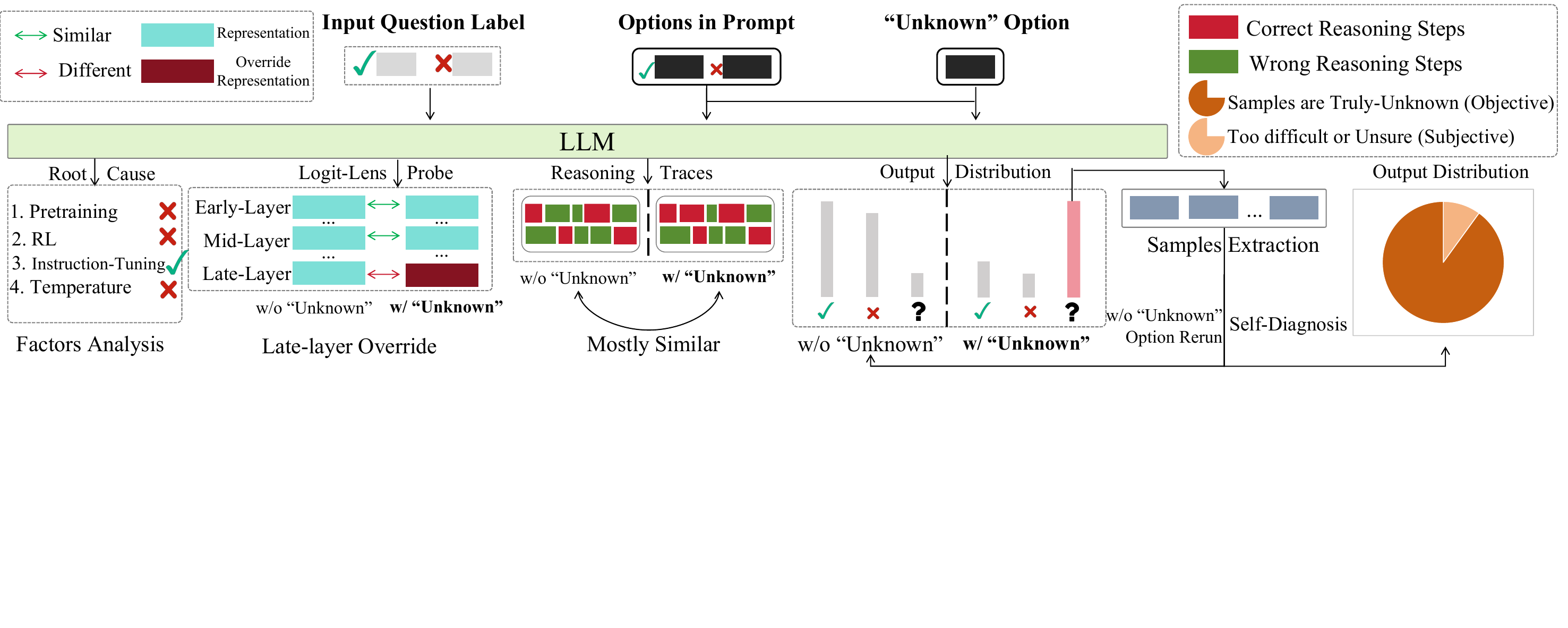}
  \vspace{-90pt}
  \caption{Representative experiments for the four claims about \emph{Abstention Inflation}, traced from prompt to training: extra-option trigger (C1), introspective gap (C2), later-layer output override (C3), various factors: instruction tuning, difficulty and scale (C4).}
    \vspace{-10pt}
  \label{fig:Main Experiments}
\end{figure*}

\section{Methodology}
\label{sec:setup}

\subsection{Claims}

We organize the study around four claims about \emph{Abstention Inflation}, each mapped to a group of experimental settings (S1--S10):

\begin{itemize}[noitemsep,topsep=2pt]
  \item \textbf{C1:} \emph{Abstention Inflation} can be triggered by the presence of an extra option, not by genuine uncertainty. (S1--S4)
  \item \textbf{C2:} \emph{Abstention Inflation} makes the models deny they can answer, even when they can. (S5--S6)
  \item \textbf{C3:} \emph{Abstention Inflation} is a later-layer output override: the reasoning trace and mid-layer representations preserve the correct answer. (S7--S8)
  \item \textbf{C4:} \emph{Abstention Inflation} is not stochastic noise: it results from various factors, emerges through instruction tuning, is boosted by problems' higher difficulty, and can be mitigated at larger model sizes. (S9--S10)
\end{itemize}

These four claims form a progressive argument. C1 establishes the phenomenon and identifies the trigger: \emph{Abstention Inflation} comes from the extra option itself, not from the word in it. C2 further shows that \emph{Abstention Inflation} cannot be genuine uncertainty: when the extra option is removed, accuracy recovers, yet the models continue to attribute their abstention to the sample being objectively unanswerable. C3 localizes this problem: the reasoning trace and mid-layer representations preserve the correct answer, but the later layers override it with the ``Unknown'' label. C4 identifies the various factors behind this bias: it persists across multiple redraws and changes little across temperatures and option positions, so it is neither stochastic noise nor a positional bias; it emerges systematically through instruction tuning, is boosted by problems' higher difficulty, and can be mitigated at larger model sizes. Together, C1$\to$C2$\to$C3$\to$C4 trace \emph{Abstention Inflation} from the prompt, to the model's introspection, to its internal representation, and finally to its training.

The corresponding Results subsections (Sections \ref{sec:rq1}, \ref{sec:rq_introspection}, \ref{sec:rq2}, \ref{sec:rq3}) each begin with a description of the relevant settings (Sx).

\subsection{Datasets}
\label{Dataset}
We study two dataset types. (1) \textbf{True/False Questions (TFQs)} ask models to classify each conclusion as True or False given the statement; we use two benchmarks: FLD~\cite{morishita2024enhancingreasoningcapabilitiesllms} ($n$=500; sampled evenly across 1--20 steps) and FOLIO~\cite{han2024folionaturallanguagereasoning} ($n$=500). Each dataset contains 250 samples labeled True and 250 samples labeled False. (2) \textbf{Multiple-Choice Questions (MCQs)} present 4-option questions with one correct answer. We use ARC~\cite{clark2018thinksolvedquestionanswering} ($n$=500), MMLU~\cite{hendrycks2021measuringmassivemultitasklanguage} ($n$=500), LogiQA~\cite{liu2020logiqachallengedatasetmachine} ($n$=500) and MedQA~\cite{jin2020diseasedoespatienthave} ($n$=500). TFQ and MCQ experiments are carried out and reported separately. Each dataset contributes equally to the final results; we also report per-dataset breakdowns in some experiments.

For Section~\ref{sec:rq3}, experiment S9 separately uses Unknown-labeled samples, i.e., samples annotated with the gold Unknown label; we extract 300 such samples from each of FLD and FOLIO. All sampling is done with seed=42.

\subsection{Models}
We test fifteen models in total: \textbf{DeepSeek-V4-Flash}, \textbf{GPT-5.4-nano}, and \textbf{Gemini-3.1-Flash-Lite} for all experimental settings except S8 and S10; for the logit-lens probe (S8), we separately use \textbf{Olmo-3-7B} in three open-source variants: Base (pretrained only), Instruct-SFT (supervised fine-tuning on Base), and RL-Zero-General (reinforcement learning on Base), enabling representation analysis across different variants; for the Factor Analysis (S10), we add \textbf{Olmo-3-7B-Instruct} for the temperature analysis, as the temperatures of some models are not adjustable; we also use the Gemma family to jointly analyze alignment and model size. We use \textbf{Gemma-4-E2B}, \textbf{Gemma-4-E4B}, \textbf{Gemma-4-26B-A4B}, and \textbf{Gemma-4-31B}, each in both pretrained base and instruction-tuned (IT) variants. All models with adjustable temperatures are set to temperature=0.0 unless otherwise specified in specific settings.

\subsection{Experimental Settings}
The ten experimental settings S1--S10 are grouped to support four claims. C1 (S1--S4) identifies the phenomenon and ablates the influence of question format and word content. C2 (S5--S6) probes the dissociation between affected accuracy and self-diagnosis. C3 (S7--S8) localizes the override mechanism via reasoning traces and logit-lens probes. C4 (S9--S10) examines the various factors behind \emph{Abstention Inflation}: stability across repeated draws, and sensitivity to difficulty, temperature, alignment, model size and the position of the ``Unknown'' option. Each setting is described at the start of the corresponding claim.

\subsection{Metrics}
\label{Metrics}
Let $\mathcal{D}$ denote the dataset, and $f_\theta$ the model prediction. \emph{\textbf{Acc}} measures label-prediction accuracy:
\begin{equation}
  \text{Acc} = \frac{|\{x \in \mathcal{D} : f_\theta(x) = y_x\}|}{|\mathcal{D}|}
\end{equation}
\emph{\textbf{Abs Rate}} measures the fraction of samples where the model outputs the ``Unknown'' option:
\begin{equation}
  \text{Abs Rate}= \frac{|\{x \in \mathcal{D} : f_\theta(x) = \text{``Unknown''}\}|}{|\mathcal{D}|}
\end{equation}
Both metrics range from 0\% to 100\%.

\begin{table*}[t]
\centering
\footnotesize
\renewcommand{\arraystretch}{0.88}
\setlength{\tabcolsep}{3.5pt}
\resizebox{\linewidth}{!}{%
\begin{tabular}{ll cccccccc}
\toprule
\textbf{Model} & \textbf{Metric}
  & \textbf{FLD}
  & \textbf{FLD\textsubscript{MCQ}}
  & \textbf{FOLIO}
  & \textbf{FOLIO\textsubscript{MCQ}}
  & \textbf{ARC}
  & \textbf{MedQA}
  & \textbf{MMLU}
  & \textbf{LogiQA} \\
\midrule
\multirow{3}{*}{DeepSeek-V4-Flash}
  & Acc (S1) & 78.4 & 78.4 & 87.0 & 87.0 & 98.4 & 94.2 & 89.2 & 73.4 \\
  & Acc (S2) & 53.0\dn{25.4} & 51.8\dn{26.6} & 75.4\dn{11.6} & 76.2\dn{10.8} & 98.8\up{0.4} & 92.6\dn{1.6} & 88.8\dn{0.4} & 69.2\dn{4.2} \\
  & Abs Rate (S2) & 42.4 & 43.2 & 19.4 & 18.2 & 0.2 & 0.6 & 1.4 & 3.6 \\
\midrule
\multirow{3}{*}{GPT-5.4-nano}
  & Acc (S1) & 61.9 & 61.9 & 84.2 & 84.2 & 93.8 & 82.8 & 76.1 & 50.6 \\
  & Acc (S2) & 23.2\dn{38.7} & 27.4\dn{34.5} & 72.0\dn{12.2} & 71.6\dn{12.6} & 94.0\up{0.2} & 79.9\dn{2.9} & 75.5\dn{0.6} & 50.2\dn{0.4} \\
  & Abs Rate (S2) & 69.4 & 63.1 & 22.2 & 21.8 & 1.0 & 2.5 & 2.4 & 5.6 \\
\midrule
\multirow{3}{*}{\makecell[l]{Gemini-3.1-\\Flash-Lite}}
  & Acc (S1) & 76.5 & 76.5 & 88.0 & 88.0 & 98.8 & 97.1 & 90.1 & 69.7 \\
  & Acc (S2) & 54.1\dn{22.3} & 51.9\dn{24.5} & 79.2\dn{8.8} & 77.6\dn{10.4} & 98.0\dn{0.8} & 94.3\dn{2.9} & 90.1\up{0.0} & 68.1\dn{1.6} \\
  & Abs Rate (S2) & 38.6 & 38.8 & 15.4 & 16.8 & 0.2 & 1.0 & 1.8 & 7.4 \\
\bottomrule
\end{tabular}
}
\vspace{-10pt}
\caption{%
  Main results. \textbf{S1 Acc}: baseline (no ``Unknown'' Option). \textbf{S2 Acc}: Accuracy with \unknown{} in prompts; subscript = $\Delta$ from S1 (\textcolor{red}{$\blacktriangledown$}decline / \textcolor{ForestGreen}{$\blacktriangle$}gain, \%).
}
\vspace{-10pt}
\label{tab:main_results}
\end{table*}

\section{Experimental Results}
\label{sec:results}

\subsection{C1: \emph{Abstention Inflation} can be triggered by the presence of an extra option, not by genuine uncertainty}
\label{sec:rq1}
\label{sec:phenomenon}

If abstention were genuinely driven by uncertainty, adding an ``Unknown'' option would not significantly change accuracy. However, experiments show the exact opposite: this option causes serious abstention behavior, and the trigger is the extra option itself rather than the option's specific semantics. We conduct experiments under four settings to show this:
\begin{itemize}[noitemsep,topsep=2pt]
  \item \textbf{S1}: Standard prompt, original label set (True/False for TFQs; A/B/C/D for MCQs). This setting establishes baseline accuracy.
  \item \textbf{S2}: ``Unknown'' is added to the S1 prompt as an extra option.
  \item \textbf{S3}: Question format ablation. TFQ samples are converted to MCQ style (A\,=\,True, B\,=\,False, C\,=\,Unknown).
  \item \textbf{S4}: Word content ablation. The ``Unknown'' option in S2 is replaced with either a synonym or a random word.
\end{itemize}

\subsubsection{S1: Baseline}

We use the prompt without the ``Unknown'' option (True/False for TFQs; A/B/C/D for MCQs). This setting establishes each model's baseline accuracy. As shown in Table~\ref{tab:main_results}, S1 Acc on TFQs ranges from 61.9\% to 88.0\%; MCQ tasks are solved at 93.8--98.8\% on ARC and 82.8--97.1\% on MedQA. These baselines confirm that all three models perform well before the ``Unknown'' option is added.  

\subsubsection{S2: ``Unknown'' Option Added}

Adding ``Unknown'' as an extra option produces a sharp accuracy drop and frequent abstention, as shown in Table~\ref{tab:main_results}: on TFQs, accuracy drops by 19.8 percentage points on average, with \emph{Abs Rate} rising to 34.5\%, indicating that many questions the models answered correctly in S1 now receive an ``Unknown'' output instead. On MCQs, the identical manipulation produces only small fluctuations.

\subsubsection{S3: Question Format Ablation}

TFQs use a True/False label format, while MCQs use letters (A/B/C/D). We test setting \textbf{S3}: TFQ samples are converted to MCQ-style labels (A = True, B = False, C = Unknown). As shown by the results for $\textnormal{FLD}_{\textnormal{MCQ}}$ and $\textnormal{FOLIO}_{\textnormal{MCQ}}$ in Table~\ref{tab:main_results}, the format manipulation leaves \emph{Abs Rate} largely unchanged for most models. Also, the changes in accuracy are far smaller than the performance drop between S1 Acc and S2 Acc, so question format is not the root cause of \emph{Abstention Inflation}.

\begin{figure}[t]
  \centering
  \includegraphics[width=\columnwidth]{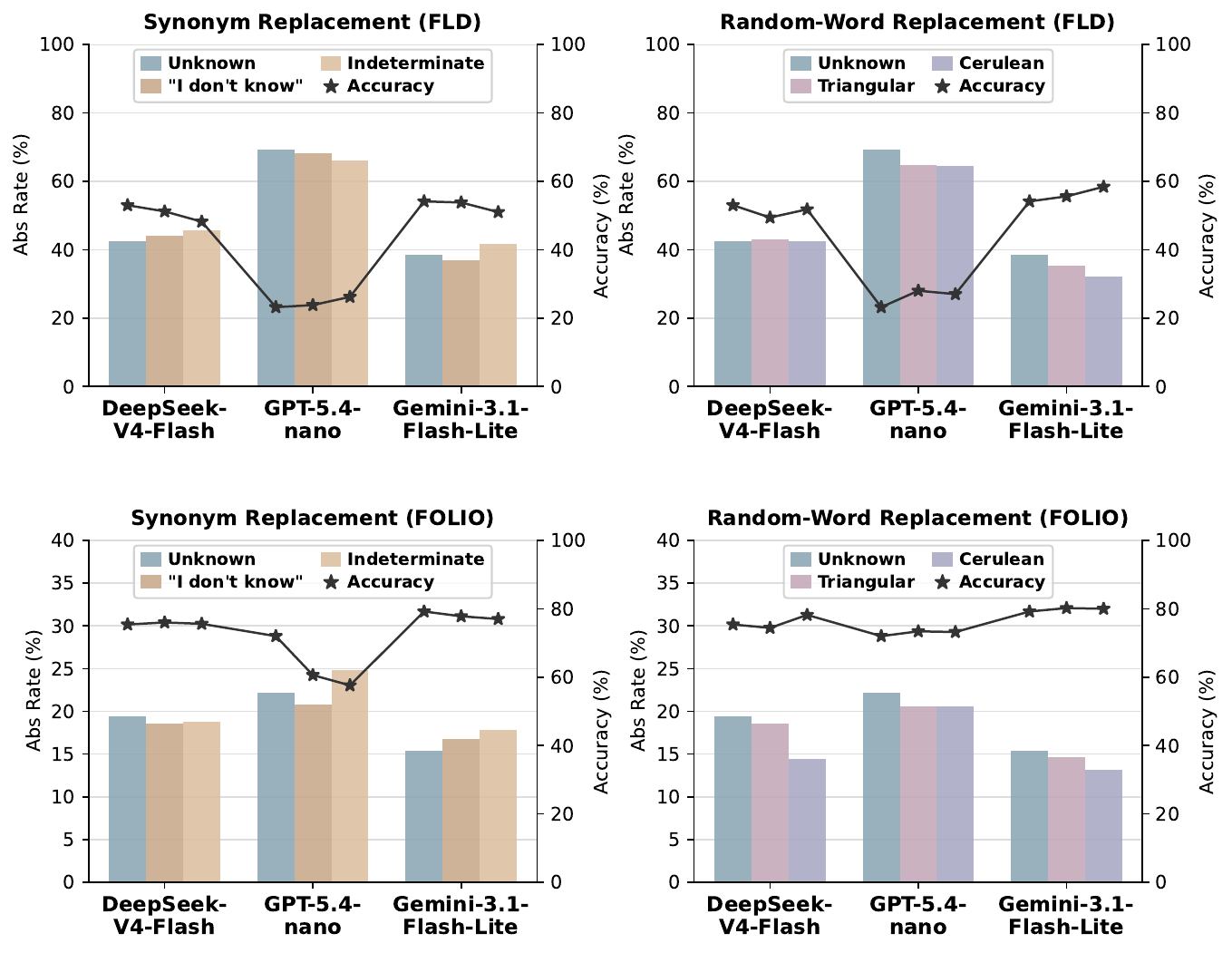}
  \caption{S4 Word Content Ablation on both TFQ datasets (top: FLD, bottom: FOLIO). Left: Replacement with synonyms of ``Unknown'' yields a similar effect. Right: Random unrelated words also yield a similar effect to ``Unknown''.}
  \vspace{-10pt}
  \label{fig:b_section}
\end{figure}

\subsubsection{S4: Word Content Ablation}
\label{sec:structure}

The results above leave open a semantic explanation: ``Unknown'' could be a trigger because models are trained on it. If so, replacing it with a less-used synonym should reduce the effect, and replacing it with a random word should eliminate it. The calculation of \emph{Abs Rate} in Section~\ref{Metrics} is also adapted to count the replacement word.

\paragraph{Synonym Replacement.}

We rerun the S2 experiment by replacing ``Unknown'' with two synonyms: ``Indeterminate'' and ``I don't know''. Results are shown on the left of Figure~\ref{fig:b_section}.

For the three variants, pairwise significance tests across all model $\times$ two datasets yield no significant differences ($\sigma \leq 2.4$\% across all experiments, all n.s.). As a result, we conclude that \emph{Abstention Inflation} is not influenced by these synonym changes.  


\paragraph{Random-Word Replacement.}

We rerun the S2 experiment across all three models on FLD and FOLIO, replacing ``Unknown'' with random words (``Triangular'' and ``Cerulean''). Results are shown on the right of Figure~\ref{fig:b_section}.

For all three models, synonyms and random words trigger abstention almost as strongly as ``Unknown'' itself, shifting \emph{Abs Rate} by at most 6.2 points. For instance, the random word ``Triangular'' triggers 30\%+ abstention on FLD and 14--21\% on FOLIO. Among responses selecting the random-word option, more than 95\% explicitly write the random word (e.g., ``Final answer: Cerulean''), confirming that the model genuinely selects it rather than treating it as noise. A representative model output reads: \textit{``\ldots I cannot determine the truth value from these premises\ldots I'll go with `Cerulean' as a placeholder for uncertainty.''} This confirms that the root cause of \emph{Abstention Inflation} is the extra option itself, rather than its specific content. More importantly, we argue that LLMs may be trained to imitate the surface pattern of \emph{abstention}, not to express genuine uncertainty.

\subsection{C2: \emph{Abstention Inflation} makes the models deny they can answer, even when they can}
\label{sec:rq_introspection}
\label{sec:c2}

Having established that \emph{Abstention Inflation} can be triggered by the presence of an extra option (C1), we ask whether the model recovers the correct answers when the ``Unknown'' option is removed. In a separate run, we test whether models can detect that this is happening at all, by evaluating the dissociation between LLMs' actual capability and their own self-diagnosis. We run two settings on the \emph{Abstention Inflation} samples from S2:
\begin{itemize}[noitemsep,topsep=2pt]
  \item \textbf{S5}: w/o ``Unknown'' Option Rerun. We rerun \emph{Abstention Inflation} samples with the ``Unknown'' option removed from the prompt to see whether the model can correctly predict the labels.
  \item \textbf{S6}: Self-Diagnosis. Models are asked to attribute their abstention to (A) subjective incapability or (B) the samples being objectively unanswerable.
\end{itemize}

\subsubsection{S5: w/o ``Unknown'' Option Rerun}

We collect \emph{Abstention Inflation} samples, i.e., samples on which the models abstain when the ``Unknown'' option is given, and rerun them with the option removed from the prompt. If models genuinely cannot answer, the accuracy should be at or below the 50\% random baseline. Instead, the average accuracy is around 66\%, well above 50\% (Figure~\ref{fig:c2_c4}, left): the models can answer correctly without the ``Unknown'' option, yet choose to abstain when this option is given. \emph{Abstention Inflation} is a bias rather than an inability.  

\subsubsection{S6: Self-Diagnosis}

\begin{figure}[t]
  \centering
  \includegraphics[width=\columnwidth]{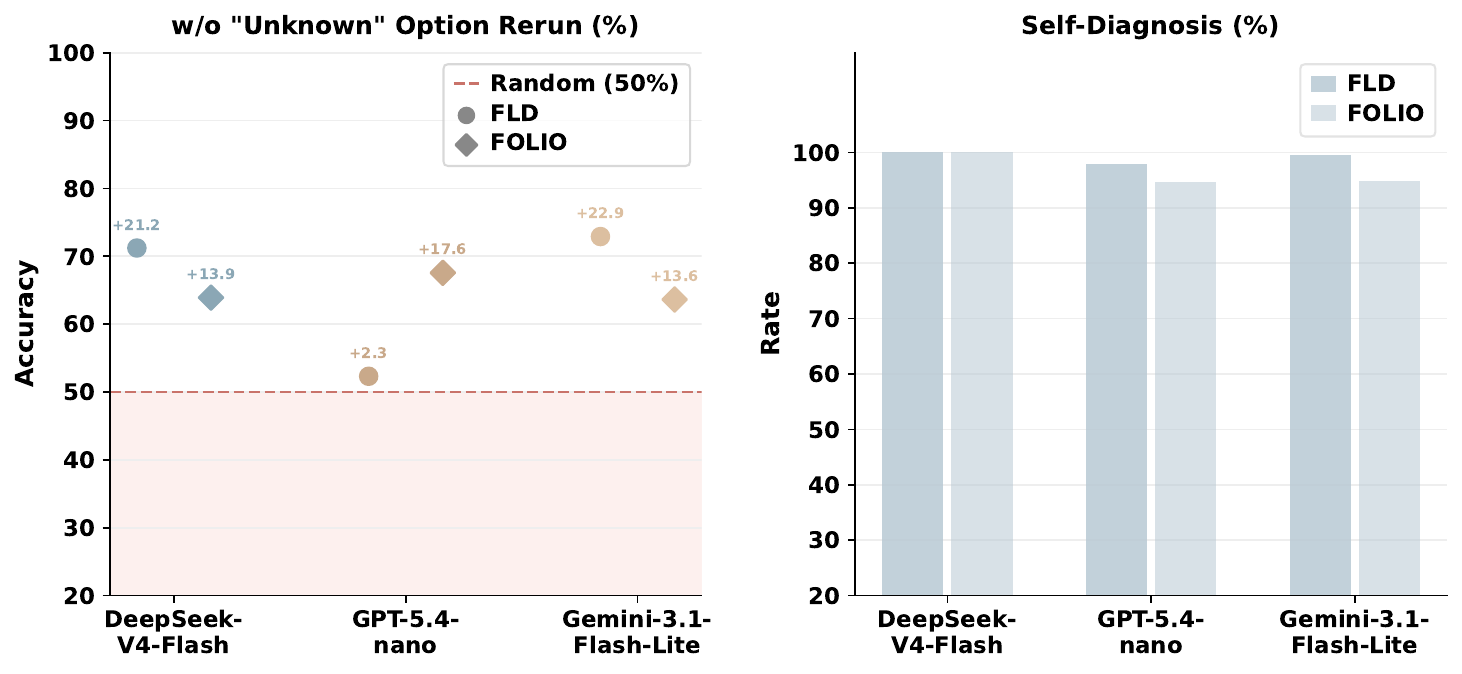}
  \caption{Left: S5, w/o ``Unknown'' Option Rerun; all accuracies are above the 50\% random baseline. Right: S6, Self-Diagnosis; models attribute 94.6--100\% of their abstentions to the samples being objectively unanswerable, despite being able to answer correctly.}
  \vspace{-10pt}
  \label{fig:c2_c4}
\end{figure}

We ask the same models to attribute their abstention: was it (A) subjective incapability or (B) the sample being objectively unanswerable? Results appear in Figure~\ref{fig:c2_c4} (right): models attribute 94.6--100\% of these abstentions to objective factors, i.e., the samples being objectively unanswerable, even though S5 just showed they could answer.  

Combining S5 and S6 reveals an introspective gap: the model can answer correctly without the ``Unknown'' option, but its own self-report denies this capability.

\subsection{C3: \emph{Abstention Inflation} is a later-layer output override: the reasoning trace and mid-layer representations preserve the correct answer}
\label{sec:rq2}
\label{sec:mechanism}

C2 establishes that the models can answer (S5) yet deny they can (S6). Also, Adding the ``Unknown'' option may also affect the model's reasoning and its representations. To evaluate this, we study where in the network the override is introduced. We show that the reasoning trace changes far less when ``Unknown'' is added, and that mid-layer representations still carry the correct answer; the problem is introduced in the later layers. To test C3, we run two settings:
\begin{itemize}[noitemsep,topsep=2pt]
  \item \textbf{S7}: A DeBERTa NLI probe reads the conclusion out of the reasoning traces generated under S1 and S2, testing whether the trace itself reaches the correct label.
  \item \textbf{S8}: A logit-lens probe tracks label-prediction probability across all layers of Olmo-3-7B variants (Base, Instruct-SFT, RL-Zero-General) for prompts with and without the ``Unknown'' option, i.e., one-word-answer variants of S1 and S2 in Section~\ref{sec:rq1} (App.~\ref{app:prompts}).
\end{itemize}

\subsubsection{S7: Reasoning Traces Evaluation}


We measure whether adding the ``Unknown'' option harms reasoning trace quality. If so, CoT quality should drop along with the accuracy drop. We quantify trace quality with a DeBERTa NLI probe that reads the conclusion out of the trace: the trace is the premise and the conclusion is the hypothesis, so \texttt{Entailment}, \texttt{Contradiction} and \texttt{Neutral} map onto True, False and Unknown labels. Since FLD/FOLIO labeling is semantically equivalent to NLI, the probe's three classes map directly onto the task's own label space.

\begin{figure}[t]
  \centering
  \includegraphics[width=\columnwidth]{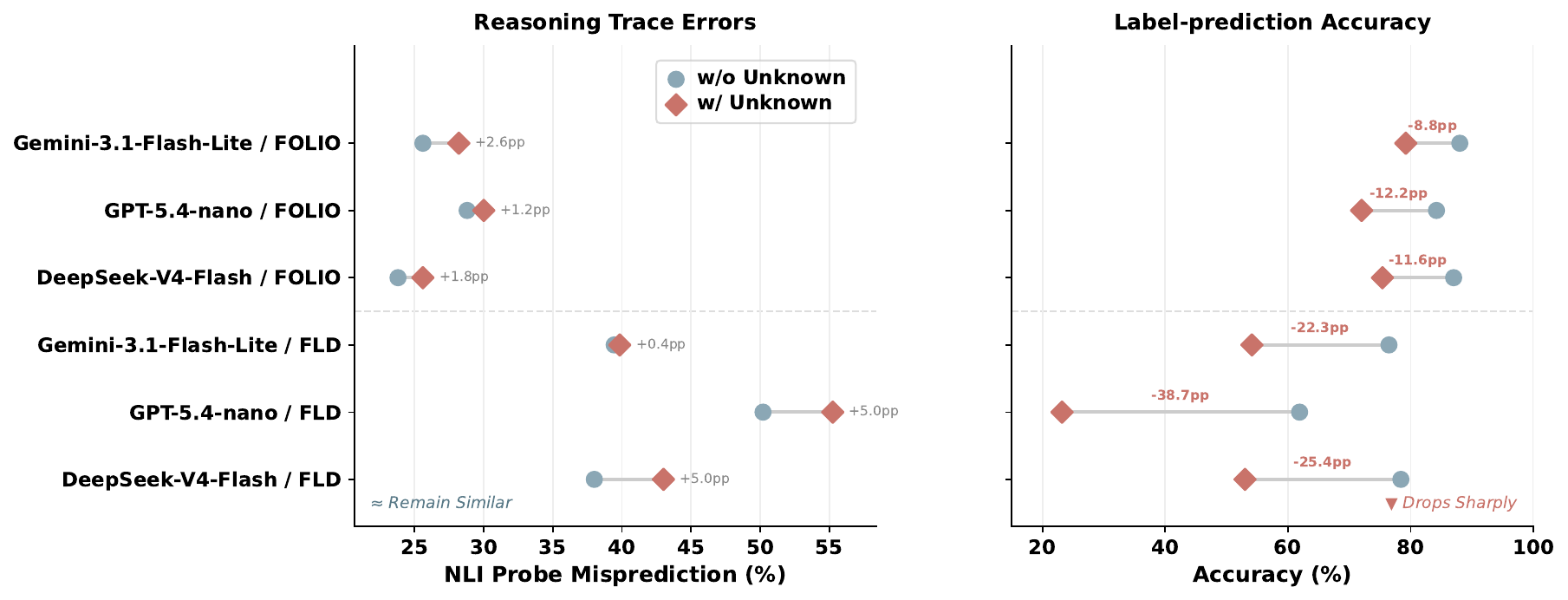}
  \caption{S7: Reasoning Traces Evaluation on traces generated with and without the ``Unknown'' option in the prompt. Left: the rate at which the NLI probe fails to recover the gold label from the trace. Right: adding the ``Unknown'' option causes a large accuracy drop.}
  \vspace{-10pt}
  \label{fig:c0}
\end{figure}

Figure~\ref{fig:c0} shows that trace quality is far less affected when ``Unknown'' is added: NLI probe misprediction rises only from 34.4\% to 37.0\% over the 500 samples, while the label-prediction accuracy falls sharply, from 79.4\% to 59.5\%.  


Among all samples on which the models exhibit \emph{Abstention Inflation}:

\noindent$\triangleright$ 40.6\% of CoTs reach the correct label, yet the output is ``Unknown''.

\noindent$\triangleright$ 37.4\% of CoTs are classified with wrong labels by the NLI probe, and the output is ``Unknown''.


Thus the models could predict the labels of these samples correctly, yet they abstain, driven by the added ``Unknown'' option. As indicated by~\citet{pavlick-kwiatkowski-2019-inherent}, NLI probes are biased toward predicting ``Neutral (Unknown)''; to ensure the validity of this evaluation, we manually verified 300 samples, and a case study is given in Appendix~\ref{CoT Case Study}.

\subsubsection{S8: Logit-Lens Representation Probe}
\label{S8: Logit-Lens Representation Probe}
\paragraph{Representation w/ and w/o ``Unknown'' Option.}
To pinpoint which layers \emph{Abstention Inflation} operates in, we run a logit-lens probe on \textbf{Olmo-3-7B} across three variants: Base (pretrained only), Instruct-SFT (instruction-tuned on Base), and RL-Zero-General (reinforcement learning on Base). We compare the last-token logits across all 33 transformer layers, with and without the ``Unknown'' option in the input prompt.

\begin{figure}[t]
  \centering
  \includegraphics[width=\columnwidth]{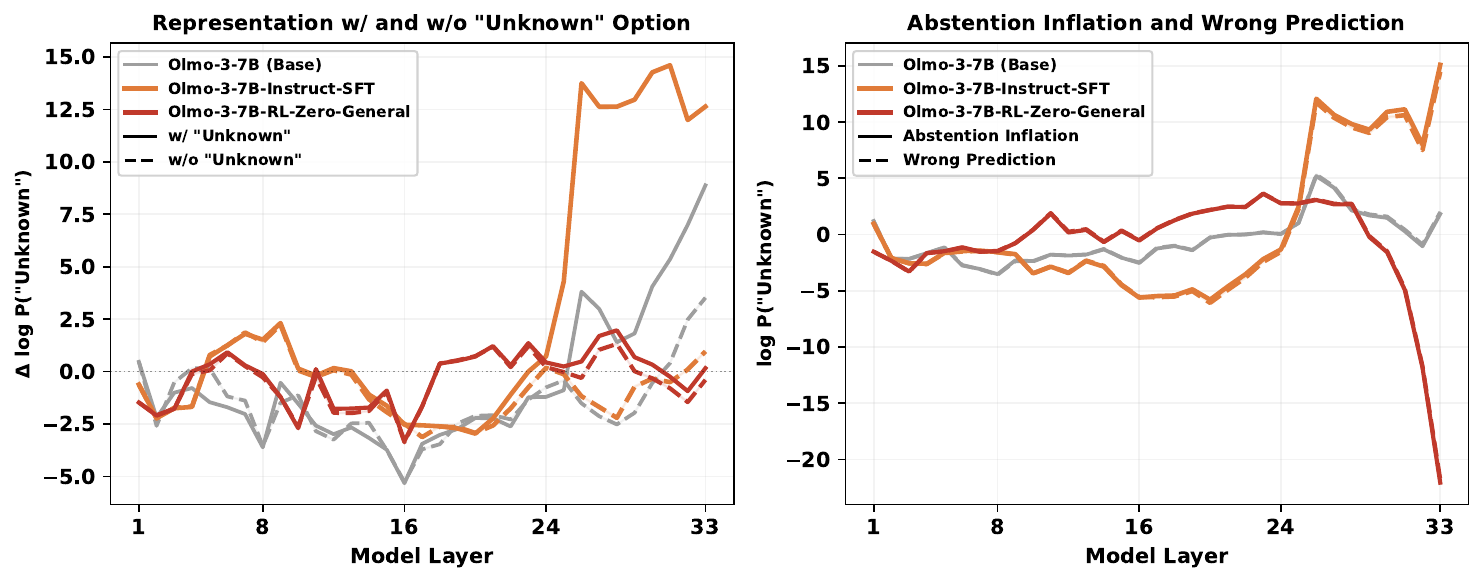}
\caption{S8: Logit-Lens Representation Probe. {Left}: the logit change $\Delta\log P(\text{``Unknown''})$ across all 33 layers for prompts with and without the ``Unknown'' option. {Right}: Comparison of $\log P(\text{``Unknown''})$ between \emph{Abstention Inflation} and \emph{Wrong Prediction}.}
\vspace{-10pt}
  \label{fig:c3}
\end{figure}

The results are shown on the left of Figure~\ref{fig:c3}. Instruct-SFT is affected the most, Base second, and RL-Zero-General the least. These results indicate that adding the ``Unknown'' option increases the corresponding token probability in the later layers, and that this effect arises mainly from SFT, i.e., supervised instruction tuning.

\paragraph{\emph{Abstention Inflation} and \emph{Wrong Prediction}.}
The right chart of Figure~\ref{fig:c3} reports a probe run on \emph{Abstention Inflation} samples, comparing them with \emph{Wrong Prediction} samples (those predicted incorrectly even without the ``Unknown'' option). Results show that the layer representation alone barely separates the two groups, as the ``Unknown'' logits in the two cases are almost equal. This is reasonable, since \emph{Abstention Inflation} also results in wrong predictions.

Experiments confirm that \emph{Abstention Inflation} (i)~is more related to instruction tuning than to other training stages; (ii)~is a later-layer output override; and (iii)~has a representation distribution nearly identical to that of \emph{Wrong Prediction} samples, making $\log P(\text{\unknown{}})$ alone insufficient for diagnosis.

\subsection{C4: \emph{Abstention Inflation} is not stochastic noise: it results from various factors, emerges through instruction tuning, is boosted by problems' higher difficulty, and can be mitigated at larger model sizes}
\label{sec:rq3}
\label{sec:bias}


C3 localized the override to the later layers of the models. We now analyze the factors behind it: whether \emph{Abstention Inflation} persists across repeated attempts and on Unknown-labeled samples, and which factors contribute to it. We support C4 with two settings:
\begin{itemize}[noitemsep,topsep=2pt]
  \item \textbf{S9}: We rerun S2 three times and test whether \emph{Abstention Inflation} is a persistent effect or a stochastic fluctuation; we also compare \emph{Abs Rate} on True/False-labeled samples and Unknown-labeled samples.
  \item \textbf{S10}: We measure different factors contributing to the  \emph{Abstention Inflation} phenomenon: task difficulty, temperature, alignment, model size, and positions of the ``Unknown'' option.
\end{itemize}

\subsubsection{S9: Stability of \emph{Abstention Inflation}}

We present two experiments in this section: the first shows that \emph{Abstention Inflation} is stable across repeated runs, ruling out random fluctuations; the second shows how the models discriminate between True/False-labeled and Unknown-labeled samples.

\begin{figure}[t]
  \centering
  \includegraphics[width=\columnwidth]{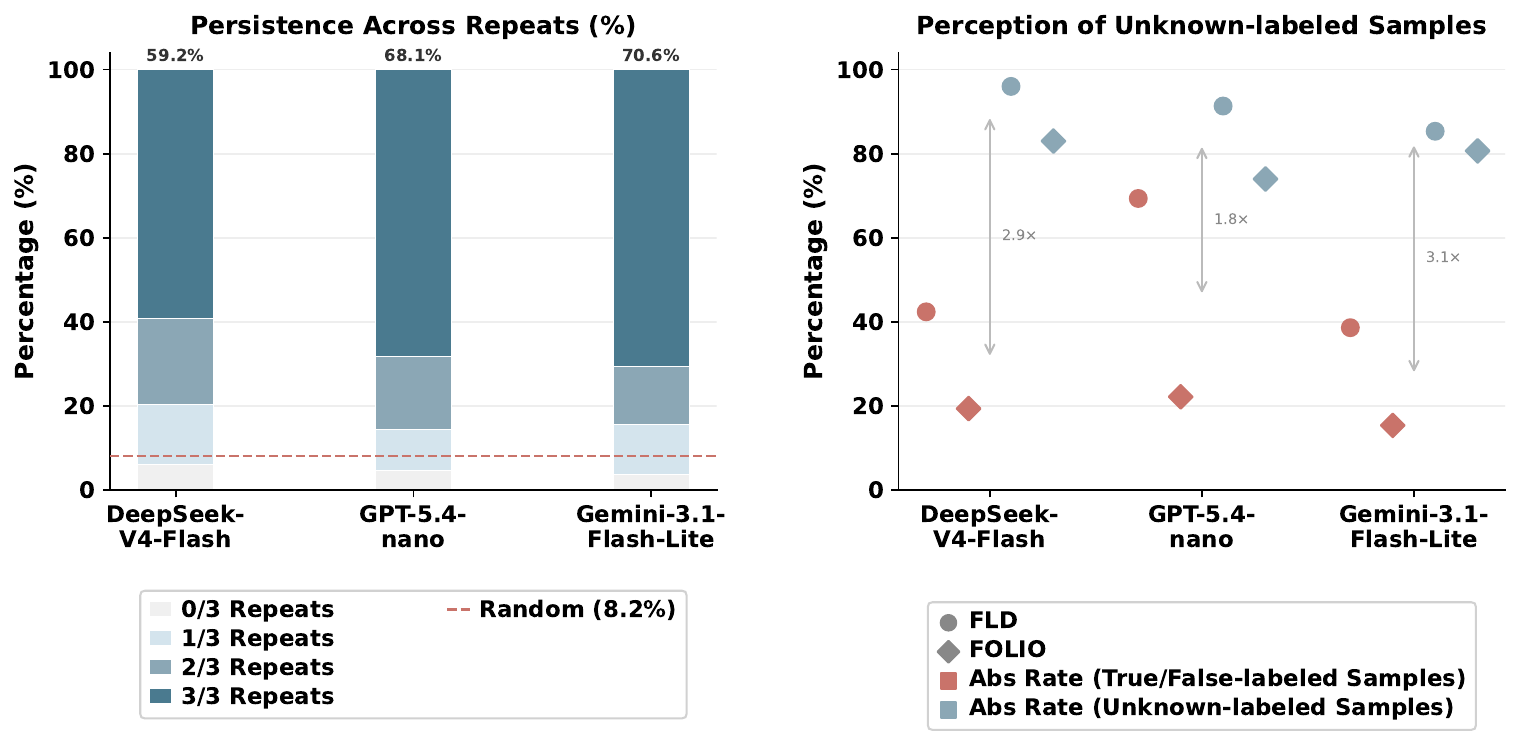}
  \caption{S9 Stability of \emph{Abstention Inflation}. Left: Persistence Across Repeats. Right: Perception of Unknown-labeled Samples.}
  \vspace{-10pt}
  \label{fig:d1_d2}
\end{figure}

\paragraph{Persistence Across Repeats.}
We rerun the experiment with the ``Unknown'' option and test whether the models persistently abstain on the same samples across multiple rounds. Figure~\ref{fig:d1_d2} (left) shows the distribution of 0/1/2/3 abstentions across repeats. Models abstain on all 3 draws for 66.1\% of \emph{Abstention Inflation} samples, which is far above the 8.2\% expected if each re-draw were an independent Bernoulli trial at the cell's own \emph{Abs Rate}. Gemini-3.1-Flash-Lite shows the most extreme persistence: 70.6\% of samples receive ``Unknown'' on all 3 rounds. We therefore conclude that \emph{Abstention Inflation} is a stable phenomenon, not a stochastic fluctuation.

\paragraph{Perception of Unknown-labeled Samples.}
As described in Section~\ref{Dataset}, we select samples from FLD and FOLIO annotated with the label ``Unknown'', i.e., Unknown-labeled samples, and run experiments with the ``Unknown'' option added. Figure~\ref{fig:d1_d2} (right) compares \emph{Abs Rate} on True/False-labeled samples with \emph{Abs Rate} on Unknown-labeled samples (where ``Unknown'' is the correct answer, so \emph{Abs Rate} equals \emph{Acc}). This shows that models are not indiscriminately selecting ``Unknown'': they correctly distinguish True/False-labeled samples from Unknown-labeled samples by a large margin. \emph{Abstention Inflation} is thus a bias toward ``Unknown'' on True/False-labeled samples, while the models still predict Unknown-labeled samples correctly.

\begin{figure}[b]
  \centering
  \includegraphics[width=\columnwidth]{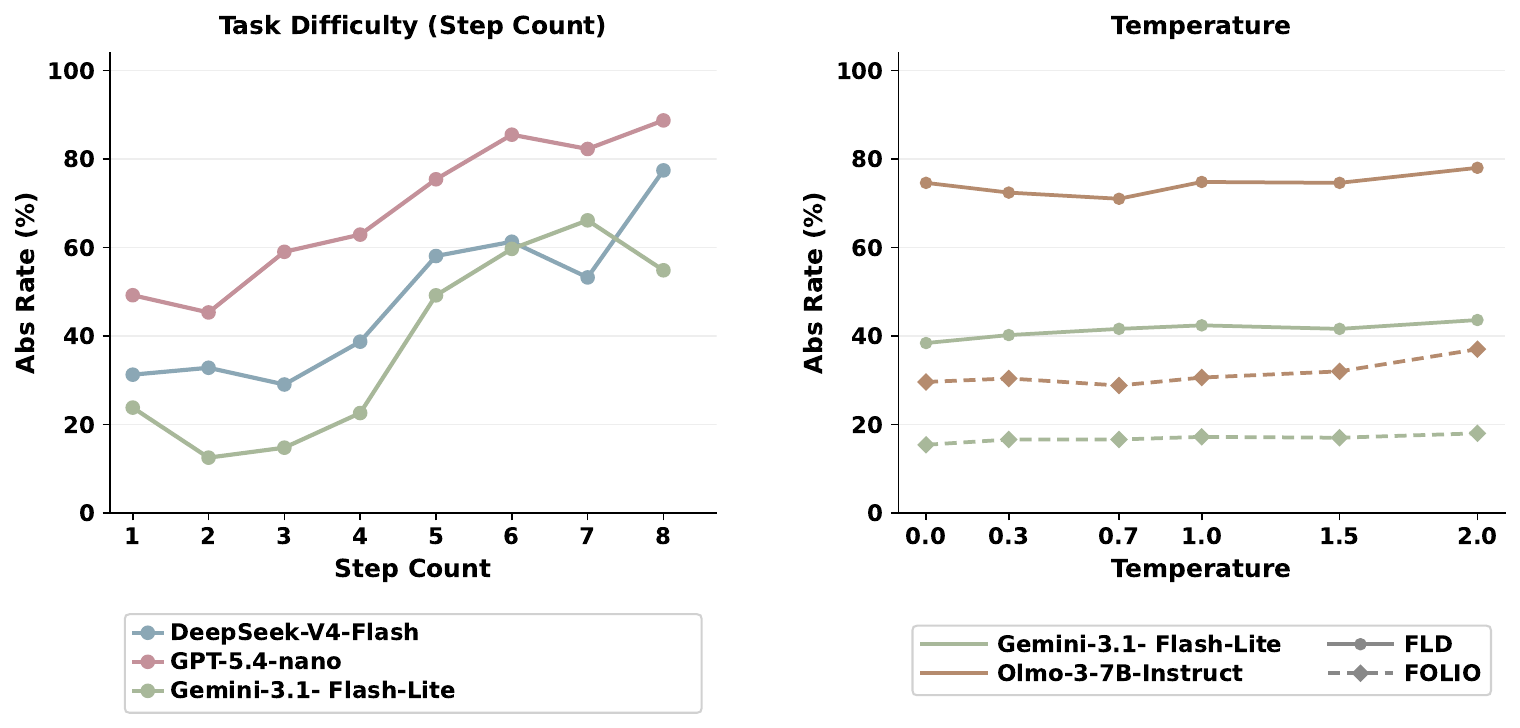}
  \caption{S10 Factor Analysis. {Left}: FLD samples with more steps lead to a higher \emph{Abs Rate}. {Right}: higher temperature raises \emph{Abs Rate} only slightly.}
  \label{fig:e1_e4}
\end{figure}

\subsubsection{S10: Factor Analysis}
\label{sec:modulators}

We examine the sensitivity of \emph{Abstention Inflation} to several factors: task difficulty (represented by step count), temperature, alignment, model size, and the position of the ``Unknown'' option. If \emph{Abstention Inflation} is a deterministic phenomenon, temperature should have little influence on it; if it activates when reasoning becomes blurry, difficulty should be positively correlated with \emph{Abs Rate}; if it is a byproduct of alignment training, base models should show markedly lower \emph{Abs Rate} than their instruction-tuned variants.

\paragraph{Task Difficulty (Step Count).}
We treat the number of steps needed to solve a problem as a measure of task difficulty. As shown on the left of Figure~\ref{fig:e1_e4}, \emph{Abs Rate} increases as the step count grows. All tested models show a stronger inclination toward abstention when problems get harder. Harder questions introduce greater inference uncertainty, thereby making the added ``Unknown'' option more likely to be chosen, but difficulty does not fully explain \emph{Abstention Inflation}: even for problems with 1--3 steps, \emph{Abs Rate} can exceed~20\%.

\begin{figure}[t]
  \centering
  \includegraphics[width=\columnwidth]{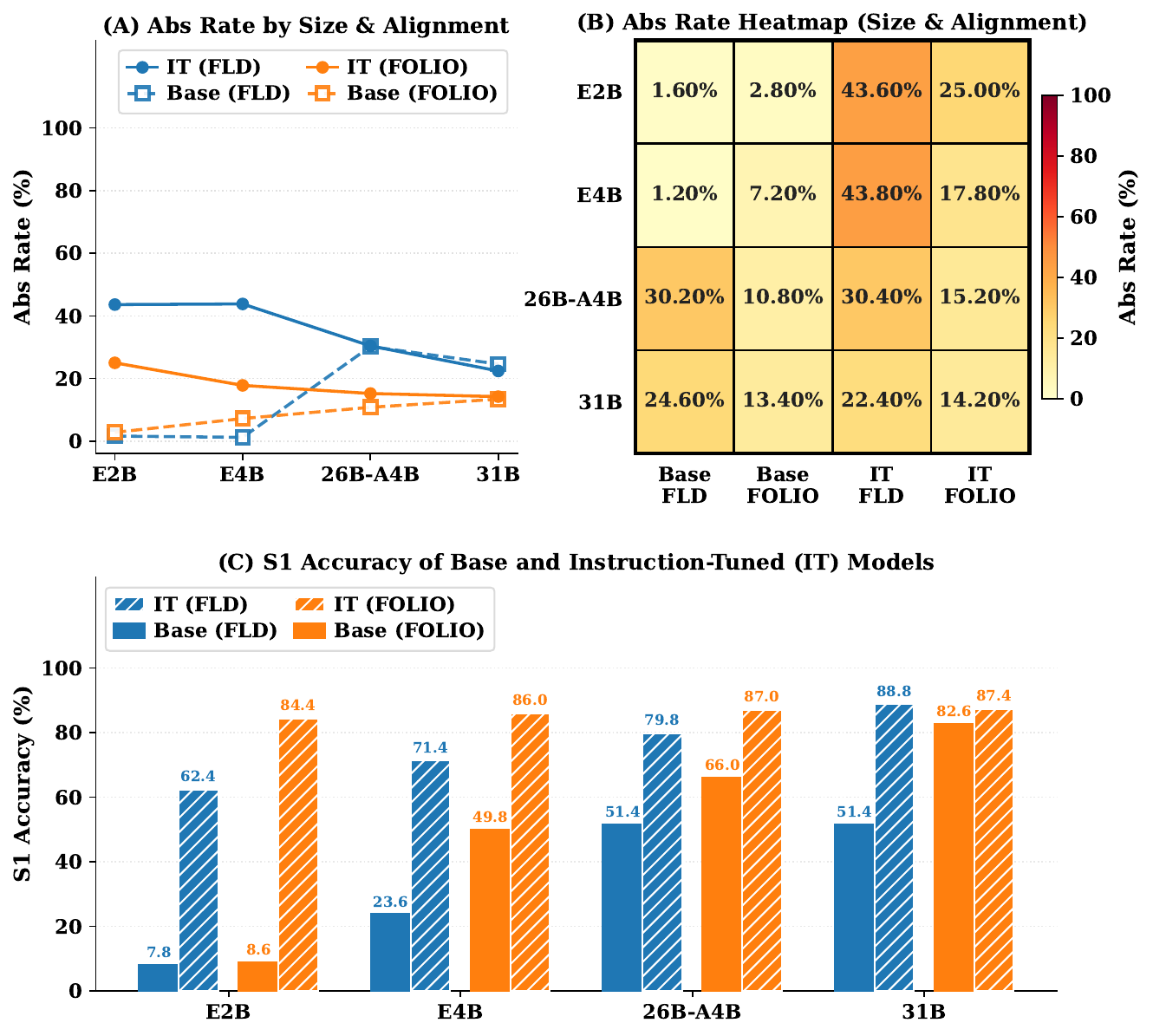}
  \vspace{-20pt}
  \caption{S10 Factor Analysis.
    \textbf{(A)}~\emph{Abs Rate} by Model Size \& Alignment.
    \textbf{(B)}~\emph{Abs Rate} heatmap.
    \textbf{(C)}~Accuracy and \emph{Abs Rate} bar chart.
  }
  \label{fig:e2}
  \vspace{-10pt}
\end{figure}

\paragraph{Temperature.}
Figure~\ref{fig:e1_e4} (right) shows \emph{Abs Rate} across $T \in \{0.0, 0.3, 0.7, 1.0, 1.5, 2.0\}$. As the temperature of some models cannot be adjusted, we add \textbf{Olmo-3-7B-Instruct} to this experiment. \emph{Abs Rate} increases slightly as temperature rises, at most 7.4 points between $T{=}0$ and $T{=}2.0$. Adjusting the temperature alone therefore cannot ease \emph{Abstention Inflation}.

\paragraph{Model Size and Alignment.}

Chart~(A) of Figure~\ref{fig:e2} separates IT and base models. As model size grows, the \emph{Abs Rate} of the two variants converges: the \emph{Abs Rate} of IT models drops as models grow, while that of base models rises. Chart~(B) gives the detailed values. Together with the findings in Section~\ref{S8: Logit-Lens Representation Probe}, we conclude that instruction tuning induces the \emph{Abstention Inflation} phenomenon and that larger model sizes mitigate the effect.



\begin{figure}[t]
  \centering
  \includegraphics[width=\columnwidth]{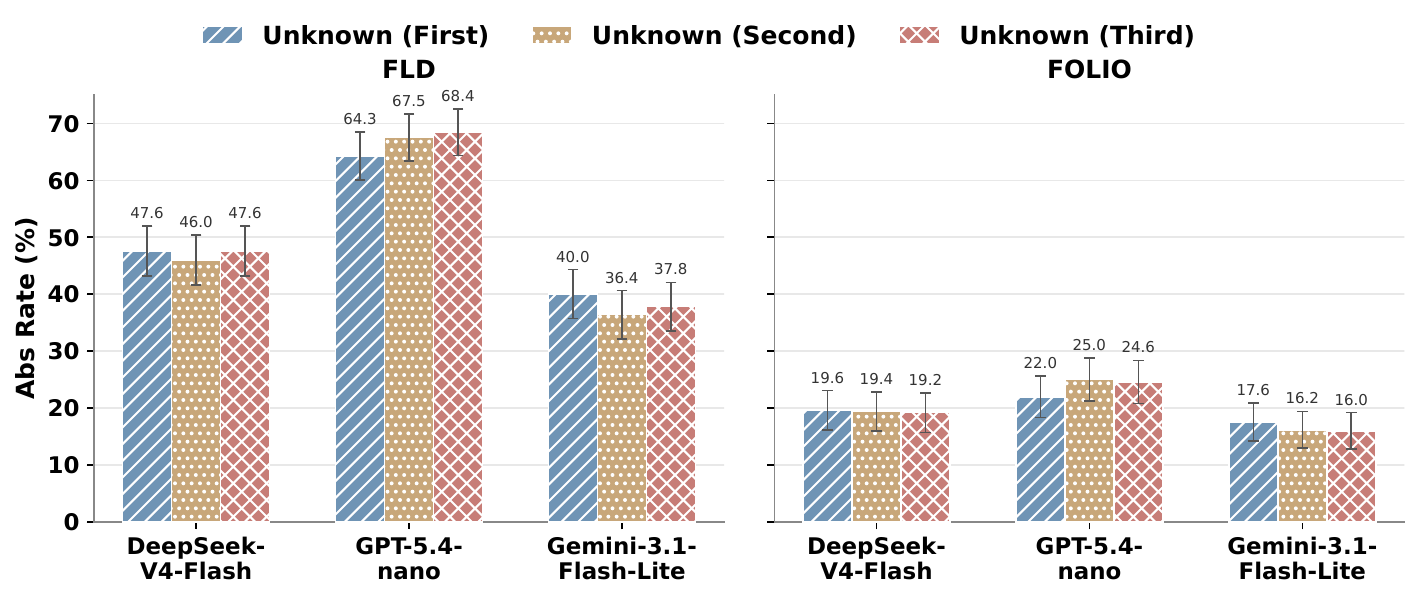}
  \caption{S10 Factor Analysis: Positional Biases. Error bars indicate 95\% Wilson confidence intervals. Unknown (First), Unknown (Second), and Unknown (Third) place the ``Unknown'' option in the first, second, and third positions, respectively.}
  \label{fig:s11}
\end{figure}

Along the alignment axis, Chart~(C) of Figure~\ref{fig:e2} shows the accuracy of base and IT models. At smaller sizes such as E2B and E4B, instruction tuning simultaneously increases accuracy and \emph{Abstention Inflation}. We attribute this to a mismatch in instruction tuning: a training example in which the model correctly abstains because the problem is objectively unanswerable (objective factor) and one in which it abstains because of difficulty (subjective factor) look identical: both end in abstention and receive the same training signal. At 26B-A4B and 31B, the two effects come apart: the accuracy benefit persists, while the gap in \emph{Abs Rate} between the two variants largely closes.

To summarize, instruction tuning contributes two effects: (i) stronger performance that raises accuracy, and (ii) activation of \emph{Abstention Inflation} whenever the models are shown an ``Unknown'' option. Both effects are stronger on smaller models. For the larger models tested, the effect is largely mitigated. \emph{Abstention Inflation} is therefore a property of instruction tuning that larger model sizes mitigate.

\paragraph{Positional Biases.}
\label{sec:s11}
To assess positional biases, we place the ``Unknown'' option in the first, second, or third position and conduct experiments on FLD and FOLIO; the third-position condition is the original S2 setting. As shown in Figure~\ref{fig:s11}, across three models and two benchmarks, \emph{Abs Rate} changes little across different positions; the largest fluctuation is 4.1 percentage points. These results indicate that \emph{Abstention Inflation} is stable and that its positional biases are weak.  

\section{Discussion}
\label{sec:discussion}

We discuss the implications and offer advice on mitigating \emph{Abstention Inflation}.





\subsection{Implications}
\label{sec:f2}

LLM abstention should be treated as a double-edged sword, rather than an always-positive signal of transparency and trustworthiness. Downstream systems such as multi-agent pipelines risk inheriting this bias as if it reflected genuine uncertainty. Future work should focus on easing the effects and potential risks of \emph{Abstention Inflation}.

\subsection{Problem Mitigation Advice}
\label{sec:f3}

\paragraph{Carefully construct prompts and implement extended reasoning.} User queries containing uncertainty can bias the model toward abstention; extended reasoning mode can help mitigate this.

\paragraph{Add verification steps to benchmark construction.} Detailed measurements should be performed when adding uncertainty categories, which may cause biases in LLMs' responses.


\section{Conclusion}
\label{sec:conclusion}

We identify and study the \emph{Abstention Inflation} phenomenon: LLM abstention may not always be a good signal of transparency and trustworthiness. Adding an extra ``Unknown'' option sharply increases abstention on TFQs across all tested models. Such abstention is unfaithful, as most questions the models abstain on are answered correctly once this option is removed. If we replace ``Unknown'' with a random word, the models treat it as a placeholder for ``Unknown'' and still choose it. The effect changes little when the position of the ``Unknown'' option changes, showing that positional biases are weak. We argue that LLM abstention can sometimes be a surface pattern, rather than a genuine expression of uncertainty. Based on our experiments, we offer advice on mitigating the problem, and our work helps in understanding LLM uncertainty expression.


\section*{Limitations}

Our experiments cover three commercial frontier models (DeepSeek-V4-Flash, GPT-5.4-nano, Gemini-3.1-Flash-Lite), two open-weight model families (Olmo-3-7B and Gemma), and six datasets (FLD, FOLIO, ARC, MMLU, MedQA, LogiQA). The \emph{Abstention Inflation} phenomenon may differ in magnitude for additional or larger model families. Our finding that the trigger is structural, not semantic, should be task-independent and generalize, but cross-model variation in the magnitude of \emph{Abstention Inflation} is an open empirical question.

\section*{Ethical Considerations}

We conduct our experiments on publicly available datasets, one of which (MedQA) is in a high-stakes domain. As our objective is to evaluate whether LLMs can appropriately abstain from answering, the introduction of an ``Unknown'' option serves to promote more conservative model behavior. We therefore believe this work raises no significant ethical concerns and is unlikely to cause societal harm.

\section*{Acknowledgment}
We thank \textbf{Yue Huang} of the University of Notre Dame for framing the idea and providing guidance, and \textbf{Guoxia Zhang} of Bosch (China) Investment Ltd. for valuable discussions and insights. This work was supported by the National Natural Science Foundation of China (Grant No.62506318); Guangdong Provincial Department of Education Project (Grant No.2024KQNCX028); Scientific Research Projects for the Higher-educational Institutions (Grant No.2024312096), Education Bureau of Guangzhou Municipality; Guangzhou-HKUST(GZ) Joint Funding Program (Grant No.2025A03J3957), Education Bureau of Guangzhou Municipality.

\bibliography{custom}

@misc{morishita2024enhancingreasoningcapabilitiesllms,
      title={Enhancing Reasoning Capabilities of LLMs via Principled Synthetic Logic Corpus}, 
      author={Terufumi Morishita and Gaku Morio and Atsuki Yamaguchi and Yasuhiro Sogawa},
      year={2024},
      eprint={2411.12498},
      archivePrefix={arXiv},
      primaryClass={cs.LG},
      url={https://arxiv.org/abs/2411.12498}, 
}

@article{pavlick-kwiatkowski-2019-inherent,
    title = "Inherent Disagreements in Human Textual Inferences",
    author = "Pavlick, Ellie  and
      Kwiatkowski, Tom",
    editor = "Lee, Lillian  and
      Johnson, Mark  and
      Roark, Brian  and
      Nenkova, Ani",
    journal = "Transactions of the Association for Computational Linguistics",
    volume = "7",
    year = "2019",
    address = "Cambridge, MA",
    publisher = "MIT Press",
    url = "https://aclanthology.org/Q19-1043/",
    doi = "10.1162/tacl_a_00293",
    pages = "677--694"
}

@misc{han2024folionaturallanguagereasoning,
      title={FOLIO: Natural Language Reasoning with First-Order Logic}, 
      author={Simeng Han and Hailey Schoelkopf and Yilun Zhao and Zhenting Qi and Martin Riddell and Wenfei Zhou and James Coady and David Peng and Yujie Qiao and Luke Benson and Lucy Sun and Alex Wardle-Solano and Hannah Szabo and Ekaterina Zubova and Matthew Burtell and Jonathan Fan and Yixin Liu and Brian Wong and Malcolm Sailor and Ansong Ni and Linyong Nan and Jungo Kasai and Tao Yu and Rui Zhang and Alexander R. Fabbri and Wojciech Kryscinski and Semih Yavuz and Ye Liu and Xi Victoria Lin and Shafiq Joty and Yingbo Zhou and Caiming Xiong and Rex Ying and Arman Cohan and Dragomir Radev},
      year={2024},
      eprint={2209.00840},
      archivePrefix={arXiv},
      primaryClass={cs.CL},
      url={https://arxiv.org/abs/2209.00840}, 
}

@misc{hou2026finsafetybenchevaluatingllmsafety,
      title={FinSafetyBench: Evaluating LLM Safety in Real-World Financial Scenarios}, 
      author={Yutao Hou and Yihan Jiang and Yuhan Xie and Jian Yang and Liwen Zhang and Hailiang Huang and Guanhua Chen and Yun Chen},
      year={2026},
      eprint={2605.00706},
      archivePrefix={arXiv},
      primaryClass={cs.CL},
      url={https://arxiv.org/abs/2605.00706}, 
}

@misc{hu2025finetuninglargelanguagemodels,
      title={Fine-tuning Large Language Models for Improving Factuality in Legal Question Answering}, 
      author={Yinghao Hu and Leilei Gan and Wenyi Xiao and Kun Kuang and Fei Wu},
      year={2025},
      eprint={2501.06521},
      archivePrefix={arXiv},
      primaryClass={cs.CL},
      url={https://arxiv.org/abs/2501.06521}, 
}

@misc{machcha2026knowingabstainmedicalllms,
      title={Knowing When to Abstain: Medical LLMs Under Clinical Uncertainty}, 
      author={Sravanthi Machcha and Sushrita Yerra and Sahil Gupta and Aishwarya Sahoo and Sharmin Sultana and Hong Yu and Zonghai Yao},
      year={2026},
      eprint={2601.12471},
      archivePrefix={arXiv},
      primaryClass={cs.CL},
      url={https://arxiv.org/abs/2601.12471}, 
}

@misc{hendrycks2021measuringmassivemultitasklanguage,
      title={Measuring Massive Multitask Language Understanding}, 
      author={Dan Hendrycks and Collin Burns and Steven Basart and Andy Zou and Mantas Mazeika and Dawn Song and Jacob Steinhardt},
      year={2021},
      eprint={2009.03300},
      archivePrefix={arXiv},
      primaryClass={cs.CY},
      url={https://arxiv.org/abs/2009.03300}, 
}

@misc{zhang2024calibratingconfidencelargelanguage,
      title={Calibrating the Confidence of Large Language Models by Eliciting Fidelity}, 
      author={Mozhi Zhang and Mianqiu Huang and Rundong Shi and Linsen Guo and Chong Peng and Peng Yan and Yaqian Zhou and Xipeng Qiu},
      year={2024},
      eprint={2404.02655},
      archivePrefix={arXiv},
      primaryClass={cs.CL},
      url={https://arxiv.org/abs/2404.02655}, 
}

@misc{ouyang2022traininglanguagemodelsfollow,
      title={Training language models to follow instructions with human feedback}, 
      author={Long Ouyang and Jeff Wu and Xu Jiang and Diogo Almeida and Carroll L. Wainwright and Pamela Mishkin and Chong Zhang and Sandhini Agarwal and Katarina Slama and Alex Ray and John Schulman and Jacob Hilton and Fraser Kelton and Luke Miller and Maddie Simens and Amanda Askell and Peter Welinder and Paul Christiano and Jan Leike and Ryan Lowe},
      year={2022},
      eprint={2203.02155},
      archivePrefix={arXiv},
      primaryClass={cs.CL},
      url={https://arxiv.org/abs/2203.02155}, 
}

@misc{clark2018thinksolvedquestionanswering,
      title={Think you have Solved Question Answering? Try ARC, the AI2 Reasoning Challenge}, 
      author={Peter Clark and Isaac Cowhey and Oren Etzioni and Tushar Khot and Ashish Sabharwal and Carissa Schoenick and Oyvind Tafjord},
      year={2018},
      eprint={1803.05457},
      archivePrefix={arXiv},
      primaryClass={cs.AI},
      url={https://arxiv.org/abs/1803.05457}, 
}

@misc{rafailov2024directpreferenceoptimizationlanguage,
      title={Direct Preference Optimization: Your Language Model is Secretly a Reward Model}, 
      author={Rafael Rafailov and Archit Sharma and Eric Mitchell and Stefano Ermon and Christopher D. Manning and Chelsea Finn},
      year={2024},
      eprint={2305.18290},
      archivePrefix={arXiv},
      primaryClass={cs.LG},
      url={https://arxiv.org/abs/2305.18290}, 
}

@misc{jin2020diseasedoespatienthave,
      title={What Disease does this Patient Have? A Large-scale Open Domain Question Answering Dataset from Medical Exams}, 
      author={Di Jin and Eileen Pan and Nassim Oufattole and Wei-Hung Weng and Hanyi Fang and Peter Szolovits},
      year={2020},
      eprint={2009.13081},
      archivePrefix={arXiv},
      primaryClass={cs.CL},
      url={https://arxiv.org/abs/2009.13081}, 
}

@misc{liu2020logiqachallengedatasetmachine,
      title={LogiQA: A Challenge Dataset for Machine Reading Comprehension with Logical Reasoning}, 
      author={Jian Liu and Leyang Cui and Hanmeng Liu and Dandan Huang and Yile Wang and Yue Zhang},
      year={2020},
      eprint={2007.08124},
      archivePrefix={arXiv},
      primaryClass={cs.CL},
      url={https://arxiv.org/abs/2007.08124}, 
}

@misc{huang2024uncertaintylanguagemodelsassessment,
      title={Uncertainty in Language Models: Assessment through Rank-Calibration}, 
      author={Xinmeng Huang and Shuo Li and Mengxin Yu and Matteo Sesia and Hamed Hassani and Insup Lee and Osbert Bastani and Edgar Dobriban},
      year={2024},
      eprint={2404.03163},
      archivePrefix={arXiv},
      primaryClass={cs.CL},
      url={https://arxiv.org/abs/2404.03163}, 
}

@inproceedings{Wen2024PerceptionKnowledgeBoundary,
  author       = {Wen, Zhihua and Tian, Zhiliang and Jian, Zexin and Huang, Zhen and Ke, Pei and Gao, Yifu and Huang, Minlie and Li, Dongsheng},
  title        = {Perception of Knowledge Boundary for Large Language Models through Semi‐open‐ended Question Answering},
  booktitle    = {Advances in Neural Information Processing Systems (NeurIPS) 37},
  year         = {2024},
  url          = {https://papers.nips.cc/paper_files/paper/2024/hash/a1e0d6fa0c30b7d4f75dd9c7ed6189f2-Abstract.html}
}

@misc{ling2026correctpredictionwrongsteps,
      title={Correct Prediction, Wrong Steps? Consensus Reasoning Knowledge Graph for Robust Chain-of-Thought Synthesis}, 
      author={Zipeng Ling and Shuliang Liu and Shenghong Fu and Yuehao Tang and Seonil Son and Yao Wan and Xuming Hu},
      year={2026},
      eprint={2604.14121},
      archivePrefix={arXiv},
      primaryClass={cs.CL},
      url={https://arxiv.org/abs/2604.14121}, 
}

@misc{li2025knowledgeboundarylargelanguage,
      title={Knowledge Boundary of Large Language Models: A Survey}, 
      author={Moxin Li and Yong Zhao and Wenxuan Zhang and Shuaiyi Li and Wenya Xie and See-Kiong Ng and Tat-Seng Chua and Yang Deng},
      year={2025},
      eprint={2412.12472},
      archivePrefix={arXiv},
      primaryClass={cs.CL},
      url={https://arxiv.org/abs/2412.12472}, 
}

@misc{christiano2023deepreinforcementlearninghuman,
      title={Deep reinforcement learning from human preferences}, 
      author={Paul Christiano and Jan Leike and Tom B. Brown and Miljan Martic and Shane Legg and Dario Amodei},
      year={2023},
      eprint={1706.03741},
      archivePrefix={arXiv},
      primaryClass={stat.ML},
      url={https://arxiv.org/abs/1706.03741}, 
}

@misc{wen2025knowlimitssurveyabstention,
      title={Know Your Limits: A Survey of Abstention in Large Language Models}, 
      author={Bingbing Wen and Jihan Yao and Shangbin Feng and Chenjun Xu and Yulia Tsvetkov and Bill Howe and Lucy Lu Wang},
      year={2025},
      eprint={2407.18418},
      archivePrefix={arXiv},
      primaryClass={cs.CL},
      url={https://arxiv.org/abs/2407.18418}, 
}

@misc{muhamed2025refusalbenchgenerativeevaluationselective,
      title={RefusalBench: Generative Evaluation of Selective Refusal in Grounded Language Models}, 
      author={Aashiq Muhamed and Leonardo F. R. Ribeiro and Markus Dreyer and Virginia Smith and Mona T. Diab},
      year={2025},
      eprint={2510.10390},
      archivePrefix={arXiv},
      primaryClass={cs.CL},
      url={https://arxiv.org/abs/2510.10390}, 
}

@misc{ren2023selfevaluationimprovesselectivegeneration,
      title={Self-Evaluation Improves Selective Generation in Large Language Models}, 
      author={Jie Ren and Yao Zhao and Tu Vu and Peter J. Liu and Balaji Lakshminarayanan},
      year={2023},
      eprint={2312.09300},
      archivePrefix={arXiv},
      primaryClass={cs.CL},
      url={https://arxiv.org/abs/2312.09300}, 
}

@misc{kirichenko2025abstentionbenchreasoningllmsfail,
      title={AbstentionBench: Reasoning LLMs Fail on Unanswerable Questions}, 
      author={Polina Kirichenko and Mark Ibrahim and Kamalika Chaudhuri and Samuel J. Bell},
      year={2025},
      eprint={2506.09038},
      archivePrefix={arXiv},
      primaryClass={cs.AI},
      url={https://arxiv.org/abs/2506.09038}, 
}

@misc{kadavath2022languagemodelsmostlyknow,
      title={Language Models (Mostly) Know What They Know}, 
      author={Saurav Kadavath and Tom Conerly and Amanda Askell and Tom Henighan and Dawn Drain and Ethan Perez and Nicholas Schiefer and Zac Hatfield-Dodds and Nova DasSarma and Eli Tran-Johnson and Scott Johnston and Sheer El-Showk and Andy Jones and Nelson Elhage and Tristan Hume and Anna Chen and Yuntao Bai and Sam Bowman and Stanislav Fort and Deep Ganguli and Danny Hernandez and Josh Jacobson and Jackson Kernion and Shauna Kravec and Liane Lovitt and Kamal Ndousse and Catherine Olsson and Sam Ringer and Dario Amodei and Tom Brown and Jack Clark and Nicholas Joseph and Ben Mann and Sam McCandlish and Chris Olah and Jared Kaplan},
      year={2022},
      eprint={2207.05221},
      archivePrefix={arXiv},
      primaryClass={cs.CL},
      url={https://arxiv.org/abs/2207.05221}, 
}

@misc{wang2026specalignefficientspecificationgroundedalignment,
      title={SpecAlign: Efficient Specification-Grounded Alignment of Large Language Models via Synthetic Data}, 
      author={Wenjie Wang and Yue Huang and Zhengqing Yuan and Han Bao and Shiyi Du and Yuchen Ma and Yue Zhao and Yanfang Ye and Xiangliang Zhang},
      year={2026},
      eprint={2606.16276},
      archivePrefix={arXiv},
      primaryClass={cs.AI},
      url={https://arxiv.org/abs/2606.16276}, 
}

\clearpage
\appendix

\section{Dataset Details}
\label{app:datasets}

\paragraph{FLD (Formal Logical Deduction).}
FLD~\cite{morishita2024enhancingreasoningcapabilitiesllms} is a synthetic formal-logic dataset in which each item consists of a set of predicate-logic facts and a single conclusion to evaluate. Gold labels are \{True, False, Unknown\}, where the Unknown label marks items whose conclusions cannot be determined from the given facts. Each item carries a proof-tree step count (the length of the minimal deduction chain). We sample 500 answerable items (250 True + 250 False; $n$=500) per experimental setting, spread evenly across step counts, and 300 Unknown-labeled items ($n$=300) for the S9 Unknown-labeled subset (matched to FOLIO's pool size for parity). For the difficulty analysis in S10, samples are grouped by their step count. Answerable (True/False-labeled) samples are used to measure \emph{Abs Rate} (the false-abstention rate); Unknown-labeled samples are used to measure accuracy on items where selecting ``Unknown'' is correct.

\paragraph{FOLIO (First-Order Logic Inference with Open-world Assumptions).}
FOLIO~\cite{han2024folionaturallanguagereasoning} is a human-annotated first-order-logic entailment dataset in which premises are expressed in natural language and conclusions are full FOL formulae. Its gold label set also includes Unknown (for conclusions neither true nor false given the premises). We use 500 answerable items (250 True + 250 False; $n$=500) per experimental setting, and 300 Unknown-labeled items ($n$=300) for the S9 Unknown-labeled subset; the second number is capped by the size of FOLIO's public Unknown-labeled pool. FOLIO's reasoning chains are shorter than FLD's and are written in natural language, which explains the systematically lower \emph{Abs Rate} across all three frontier models (15.4--22.2\% for FOLIO vs.\ 38.6--69.4\% for FLD; see Table~\ref{tab:main_results}).

\paragraph{ARC-Challenge.}
ARC-Challenge~\cite{clark2018thinksolvedquestionanswering} is a 4-option science MCQ benchmark requiring knowledge-grounded reasoning. All items have a single correct answer; there is no Unknown-labeled subset. We use 500 items ($n$=500) per model, sampled from the challenge partition. Baseline accuracy (S1) ranges from 93.8\% to 98.8\% across our three frontier models, so these items are well within the models' knowledge.

\paragraph{MedQA.}
MedQA~\cite{jin2020diseasedoespatienthave} is a 4-option medical licensing exam benchmark. All items are answerable; we use 500 items ($n$=500) per model. Baseline S1 accuracy ranges from 82.8\% (GPT-5.4-nano) to 97.1\% (Gemini-3.1-Flash-Lite).

\paragraph{MMLU.}
MMLU~\cite{hendrycks2021measuringmassivemultitasklanguage} is a 4-option benchmark covering 57 subjects across STEM, humanities, social sciences, and other domains. All items are answerable. We use 500 items ($n$=500) per model, sampled across subjects with seed=42. Baseline S1 accuracy ranges from 76.1\% (GPT-5.4-nano) to 90.1\% (Gemini-3.1-Flash-Lite).

\paragraph{LogiQA.}
LogiQA~\cite{liu2020logiqachallengedatasetmachine} is a 4-option logical-reasoning MCQ benchmark sourced from Chinese civil-service examinations and translated to English; items require multi-step deductive or analogical inference over short natural-language passages. All items are answerable (no Unknown-labeled subset). We use 500 items ($n$=500) per model, sampled uniformly with seed=42. Baseline S1 accuracy is lower than that on the other MCQ benchmarks (50.6--73.4\% across our three models), making it a useful stress test for whether \emph{Abstention Inflation} interacts with task difficulty on MCQ-format inputs.  

\section{Experimental Settings and Evaluation Details}
\label{app:settings}

\paragraph{Setting definitions.}
\begin{itemize}[noitemsep,topsep=2pt]
  \item \textbf{S1 (Baseline)}: Standard prompt with the original label set (True/False for TFQs; A/B/C/D for MCQs). No ``Unknown'' option. This setting provides the baseline accuracy.
  \item \textbf{S2 (``Unknown'' Option Added)}: ``Unknown'' is added as an explicit additional option. For TFQs: True/False/Unknown. For MCQs: A/B/C/D/``Unknown'' (fifth option). This setting measures \emph{Abs Rate} on answerable items and accuracy on Unknown-labeled items.
  \item \textbf{S3 (Question Format Ablation)}: TFQ samples are re-rendered with MCQ-style letter labels (A\,=\,True, B\,=\,False, C\,=\,Unknown). The task instruction, context block, claim block and CoT instruction are byte-for-byte identical to S2, so question format is the only manipulation. This setting tests whether surface question format drives \emph{Abstention Inflation}.
  \item \textbf{S4 (Word Content Ablation)}: ``Unknown'' is replaced with synonyms (``Indeterminate'', ``I don't know'') and random words (``Triangular'', ``Cerulean''). This setting tests whether the ``Unknown''-option effect is semantic or structural.
  \item \textbf{S5 (w/o ``Unknown'' Option Rerun)}: Abstaining samples are re-run with the ``Unknown'' option removed; the model must commit to True or False. This setting tests whether the answer is recoverable.
  \item \textbf{S6 (Self-Diagnosis)}: The model is asked to attribute its abstention to (A) subjective incapability or (B) the sample being objectively unanswerable. This setting tests whether the model's self-report aligns with its actual capability.
  \item \textbf{S7 (Reasoning Traces Evaluation)}: Reasoning traces generated under S1 and S2 are compared with a DeBERTa NLI probe that reads each trace's own conclusion. This setting tests whether \emph{Abstention Inflation} degrades chain-of-thought quality.
  \item \textbf{S8 (Logit-Lens Representation Probe)}: Olmo-3-7B (Base, Instruct-SFT, RL-Zero-General) is run on FLD under both S1 and S2. The hidden state of each of the 33 model layers is projected through the language head; we track $\Delta\log P(\text{\unknown{}})$ across layers (Figure~\ref{fig:c3}, left: format trigger) and compare $\log P(\text{\unknown{}})$ under S2 between \emph{Abstention Inflation} samples and \emph{Wrong Prediction} samples (Figure~\ref{fig:c3}, right).
  \item \textbf{S9 (Stability)}: Each \emph{Abstention Inflation} sample is resampled three times at the default temperature. This setting tests whether abstention is a stable bias or a stochastic fluctuation.
  \item \textbf{S10 (Factor Analysis)}: Four groups of modulators are measured: (a)~temperature sweep at $T \in \{0, 0.3, 0.7, 1.0, 1.5, 2.0\}$ on FLD and FOLIO, restricted to the models whose sampling temperature is adjustable, with Olmo-3-7B-Instruct added for this sweep (tests stochasticity); (b)~\emph{Abs Rate} vs.\ FLD proof-step count (tests difficulty); (c)~an open-weight family at four scales in base and IT form on FLD and FOLIO (traces the alignment origin of \emph{Abstention Inflation}); (d)~positional biases: the TFQ version of S2 is rerun with the ``Unknown'' option in the first, second, or third position on FLD and FOLIO using the three frontier models, where the third-position condition reproduces the original S2 ordering (Section~\ref{sec:s11}).
\end{itemize}

\paragraph{Per-cell significance testing.}
Significance testing for the per-cell results below uses McNemar's test on matched item-level binary outcomes (abstain/not-abstain) for each model $\times$ dataset cell.

\paragraph{S4 Word Content Ablation: per-cell \emph{Abs Rate}.}
Per-model, per-dataset \emph{Abs Rate} under the S4 Word Content Ablation variants (``Triangular'' and ``Cerulean'') stays within 6.2 points of the S2 baseline across all six cells (mean $|\Delta|$ of 2.5 points), so random words do not change abstention behavior.

\paragraph{A. C1 (S1--S4).}
The TFQ vs.\ MCQ asymmetry is the largest single effect in the paper: when results are pooled across 3 models $\times$ 2 TFQ datasets at 500 items each ($n{=}2{,}993$ paired items; 7 of the 3{,}000 returned no usable response in one setting and are dropped from the paired tests), adding ``Unknown'' drops accuracy by 19.8 percentage points on average (McNemar $\chi^2$ values per cell range from 34.2 (Gemini-3.1-Flash-Lite $\times$ FOLIO) to 158.6 (GPT-5.4-nano $\times$ FLD)); the matched test on the 12 MCQ cells (ARC, MedQA, MMLU, LogiQA across 3 models, 500 items each; $n{=}5{,}965$ paired) gives $\chi^2{=}13.1$ for a drop of 1.2 points. S3 moves \emph{Abs Rate} by at most 6.3 points in any of the 6 cells, with a pooled mean $|\Delta\text{Abs Rate}|$ of 1.7 points. That is an order of magnitude smaller than the 34.5\% S1$\to$S2 jump, so format alone cannot account for \emph{Abstention Inflation}. Per-cell paired McNemar tests against S2 are n.s.\ in 5 of the 6 cells; GPT-5.4-nano $\times$ FLD is the exception at $p{=}0.002$, which still does not clear the corrected threshold of the 30-test structural-null family, and its \emph{Abs Rate} moves only 6.3 points. The claim rests on these magnitudes. S4 covers 4 substitution variants $\times$ 6 cells (24 tests). All 12 synonym tests are n.s., consistent with the standard deviation of at most 2.4 points reported in Section~\ref{sec:structure}. Three of the 12 random-word tests fall below $p{=}0.05$, all with ``Cerulean'' and all toward fewer abstentions than ``Unknown'' (DeepSeek-V4-Flash $\times$ FOLIO, $-5.0$ points, $p{=}0.0002$; Gemini-3.1-Flash-Lite $\times$ FLD, $-6.2$ points, $p{=}0.0013$; GPT-5.4-nano $\times$ FLD, $-4.2$ points, $p{=}0.042$). Random words therefore keep most of the effect of ``Unknown'': the largest shift is 6.2 points and the mean is 2.5.  

\paragraph{B. C2 (S5--S6).}
S5's per-cell w/o ``Unknown'' Option Rerun rates range from 52.5\% to 83.4\%; four of the six cells exceed 50\% at $p<0.01$, the exceptions being GPT-5.4-nano $\times$ FLD (52.5\%, $p{=}0.39$) and Gemini-3.1-Flash-Lite $\times$ FOLIO (63.6\%, $p{=}0.022$); the pooled estimate is 66.1\%, far above chance ($p{<}10^{-17}$). S6's self-diagnosis attributes 94.6--100\% of abstentions to the samples being objectively unanswerable across every (model, dataset) cell; a one-sample binomial test against 50\% gives a pooled $p{<}10^{-10}$. Together, S5 and S6 establish a dissociation between capability and self-report: the model can answer, yet says it cannot.  

\paragraph{C. C3 (S7--S8).}
S7's trace-level result is the central later-layer signature. Pooled over the six cells of 500 answerable items, the NLI probe fails to read the gold label out of 34.4\% of S1 traces and 37.0\% of S2 traces; the paired McNemar test detects that 2.6-point shift ($p{=}0.0015$), but it is an order of magnitude smaller than the 19.8-point accuracy drop on the same items. On the abstaining samples the trace degrades more (49.7\% to 59.4\% error, McNemar $p{<}10^{-8}$), and among the S2 traces that commit either way only 52.0\% point at the gold label ($n{=}805$, binomial $p{=}0.26$), against 58.2\% for the S1 traces of the same items ($n{=}893$, $p{<}10^{-6}$). The trace readout alone therefore does not establish that the answer survived to the output layer; S5 does. S8's per-layer paired $t$-test on the per-item $\Delta\log P(\unknown)$ between S1 and S2 prompts isolates the later-layer effect by magnitude rather than by significance: averaged over layers 26--33 the shift is $+13.9$ for Instruct-SFT, $+4.7$ for Base and $+0.63$ for RL-Zero-General ($n{=}800$ paired items per checkpoint, every one $p{<}10^{-40}$). At this sample size even RL-Zero-General's shift is detectable, so the claim rests on the 22-fold gap between Instruct-SFT and RL-Zero-General rather than on a null result for RL-Zero-General. The right panel of Figure~\ref{fig:c3} corresponds to a two-sample $t$-test between \emph{Abstention Inflation} and \emph{Wrong Prediction} groups on $\log P(\unknown)$ under S2 at layer 33; the 75 items that fall in both groups are left out of the test, leaving $n{=}95$ and $n{=}111$. The gap is $1.0$ logit for Instruct-SFT ($p{<}10^{-26}$), $0.3$ for Base ($p{=}0.01$) and $0.5$ for RL-Zero-General ($p{<}10^{-3}$): the two failure modes are statistically separable, but they sit within one logit of each other against S1$\to$S2 shifts of $+11.7$ and $+5.4$ for the first two checkpoints.  

\paragraph{D. C4a: Stability and discriminability (S9).}
The two-proportion $z$ test between \emph{Abs Rate} on answerable items (34.5\%) and on Unknown-labeled items (85.1\%) yields $z{=}34.0$, $p{<}10^{-12}$ at the pooled scale; per-cell tests are significant at $p<0.001$ for every (model, dataset). The 3/3 persistence rate is tested against the Bernoulli baseline $p_0^3$, where $p_0$ is that cell's own S2 \emph{Abs Rate}, i.e.\ the rate expected if each re-draw were an independent draw from the same abstention propensity (pooled $p_0^3{=}8.2\%$); the observed 66.1\% pooled rate is $\sim$8.1$\times$ this baseline, binomial $p{<}10^{-10}$. Gemini-3.1-Flash-Lite is the most extreme model (70.6\% against the same baseline, $p{<}10^{-8}$).  

\paragraph{E. C4b: Modulators (S10).}
Spearman $\rho$ between FLD proof-step count and \emph{Abs Rate} yields $+0.32$ ($p{<}10^{-30}$ pooled across the three models); per-model tests give $\rho_\text{DS}{=}+0.30$, $\rho_\text{GPT}{=}+0.33$, $\rho_\text{Gem}{=}+0.36$ (all $p<10^{-10}$). The temperature sweep is the weakest modulator: one-way ANOVA across $T\in\{0,0.3,0.7,1.0,1.5,2.0\}$ is n.s.\ in every swept (model, dataset) cell ($F$ between 0.26 and 2.02, all $p>0.07$). The paired $T{=}0$ vs.\ $T{=}2.0$ McNemar test is more sensitive and does detect a small upward drift in 2 of the 4 cells (Gemini-3.1-Flash-Lite $\times$ FLD, $+5.2$ points, $p{=}0.014$; Olmo-3-7B-Instruct $\times$ FOLIO, $+7.4$ points, $p{<}0.001$), so temperature is not strictly inert, but even the largest drift is about a fifth of the 34.5-point S1$\to$S2 jump in \emph{Abs Rate}, and raising $T$ moves \emph{Abs Rate} up rather than down. The alignment test paired Base vs.\ IT \emph{Abs Rate} across (4 sizes $\times$ 2 datasets) $=$ 8 matched cells is significant pooled ($p<0.01$), with the lift concentrated at E2B and E4B and close to zero at 26B-A4B and 31B (\S\ref{sec:modulators}); the same paired comparison on Acc yields a lift of comparable magnitude (App.~\ref{app:more:tax}: $\tau \approx 0.8$--$1.5$ at the smaller scales), so Base$\to$IT raises both capability and \emph{Abs Rate}. This decoupling is what grounds the attribution to SFT (App.~\ref{app:f2-details}).

\paragraph{F. C4c: Positional Biases (S10).}
Across the three frontier models and two TFQ datasets, moving the ``Unknown'' option among the first, second, and third positions changes \emph{Abs Rate} only modestly. The largest within-cell range is 4.1 percentage points, and the third-position condition reproduces the original S2 values. Figure~\ref{fig:s11} reports 95\% Wilson confidence intervals. We report the positional analysis descriptively and do not include it in the multiple-comparison family below.  

\paragraph{Multiple-comparisons note.}
Because the paper reports many tests, we provide Holm--Bonferroni-adjusted thresholds for the inferential comparisons in Groups A--E; Group F is descriptive. With 24 S4 tests and 6 S3 tests for the structural-null claim (Group A), the Holm threshold starts at $0.05/30 \approx 0.0017$. The two ``Cerulean'' cells at $p{=}0.0002$ and $p{=}0.0013$ clear it, so for those two cells the random word is a slightly weaker trigger than ``Unknown''; the next smallest $p$ (the S3 cell at $p{=}0.002$) does not. Both surviving shifts are 6.2 points or less, small next to the 34.5-point S1$\to$S2 jump. Among the pooled tests, every significant effect reported above clears $\alpha{=}0.001$ before correction except the S7 trace shift ($p{=}0.0015$); the per-cell exceptions (S5 Gemini-3.1-Flash-Lite $\times$ FOLIO and the two temperature-drift cells) are already flagged individually. Apart from these two S4 cells, multiple-comparisons correction does not change any reported decision.

\section{Additional Quantitative Analyses}
\label{app:more}

This section collects supplementary analyses of the existing experimental data; none of them needs new API calls. Each subsection takes one statement that the main results imply and puts a number on it.

\subsection{Abstention Rivals Baseline Accuracy}
\label{app:more:evidence}

GPT-5.4-nano on FLD (Table~\ref{tab:main_results}) is the clearest single case against the genuine-uncertainty hypothesis. Its S1~Acc is 61.9\% and its \emph{Abs Rate} is 69.4\%, so once the option is offered it abstains on more answerable items than it answered correctly without it. DeepSeek-V4-Flash on FLD shows the same thing from the other side: it reaches 78.4\% S1~Acc and still abstains on 42.4\% of items. Epistemic humility explains neither case; both are \emph{Abstention Inflation}. Together with S5's pooled w/o ``Unknown'' Option Rerun recovery of $\sim$66\% on \emph{Abstention Inflation} samples (\S\ref{sec:c2}), these cells show that \emph{Abstention Inflation} operates independently of what the model actually knows.  

\subsection{Dataset Modulation of \emph{Abstention Inflation}}

\emph{Abs Rate} is systematically higher on FLD (38.6--69.4\% across the three frontier models) than on FOLIO (15.4--22.2\%; Table~\ref{tab:main_results}), and the Base$\to$IT \emph{Abs Rate} lift is correspondingly larger on FLD. FLD's multi-step formal deduction chains make the inference boundary fuzzier, in line with S10's task-difficulty effect ($\rho = +0.32$, \S\ref{sec:modulators}), while FOLIO's natural-language premises hold the model more firmly to a committed answer. Alignment installs \emph{Abstention Inflation}, and task structure then sets how strongly it shows. This interaction is second order and does not undermine the primary claim.

\subsection{Abs Rate--\texorpdfstring{$\Delta$}{Δ}Acc Linear Regression}
\label{app:regression}

As a post-hoc analysis on the existing outputs, we fit OLS regressions between \emph{Abs Rate} and $\Delta\text{Acc}$ across all 18 model $\times$ dataset cells (6 TFQ, 12 MCQ) of Table~\ref{tab:main_results}.

\paragraph{Full-sample results.}
Pearson $r = -0.992$, $R^2 = 0.984$, $p = 1.1 \times 10^{-15}$ (Spearman $\rho = -0.807$). Higher \emph{Abs Rate} strongly predicts larger accuracy degradation.

\paragraph{TFQ subset.}
Pearson $r = -0.998$, $R^2 = 0.996$, $p = 5.6 \times 10^{-6}$ (Spearman $\rho = -1.000$). OLS slope $= -0.56$: a 10-point increase in \emph{Abs Rate} predicts a 5.6-point accuracy loss. The relationship is nearly perfectly linear within TFQs.

\paragraph{MCQ subset.}
The correlation is not significant ($r = -0.229$, $p = 0.47$, n.s.): because \emph{Abs Rate} $\approx 0$ and $\Delta\text{Acc} \approx 0$ for MCQs, there is little variance to correlate.

The TFQ and MCQ regression lines are completely separated, which is another view of the double dissociation. For practitioners deploying TFQ-format tasks with an ``Unknown'' option, the regression gives a quantitative estimate: the observed \emph{Abs Rate} forecasts the accuracy cost.

\subsection{Decomposing S5 w/o ``Unknown'' Option Rerun into a ``Known'' Share}
\label{app:more:decomp}

A w/o ``Unknown'' Option Rerun probe on an \emph{Abstention Inflation} sample succeeds either because the model truly knows the answer (it picks the correct option deterministically) or because it guesses between True and False (and is correct with probability $0.5$). Let $\alpha$ be the share of \emph{Abstention Inflation} samples on which the model truly knows the answer. Then
\[
P(\text{correct} \mid \text{forced}) \;=\; \alpha \cdot 1 + (1-\alpha)\cdot\tfrac{1}{2},
\]
\[
\alpha \;=\; 2P_{\text{forced}} - 1
\]
Applying this to the per-cell S5 w/o ``Unknown'' Option Rerun accuracies (52.5--83.4\% across the six model $\times$ dataset cells) yields a per-cell ``known'' share $\alpha$ between roughly 5\% and 67\%.  

The pooled $\bar\alpha \approx 32\%$ is a strict lower bound on the share of abstentions that are \emph{Abstention Inflation}, and it is of the same order as the 40.6\% of those traces from which the S7 probe reads the gold label. The two numbers do not carry equal weight. On this subset the probe's verdict cannot be distinguished from its opposite (52.0\% of committed traces, $p{=}0.26$), so the mechanism claim rests on the behavioral estimate.  

\subsection{Effect Hierarchy: Structural vs.\ Surface Modulators}
\label{app:more:corr}

Collecting the ablations of Sections~\ref{sec:rq1} and~\ref{sec:modulators} in one place orders them by effect size: question type (TFQ vs.\ 4-option MCQ) and alignment (Base vs.\ IT) move \emph{Abs Rate} by tens of points, whereas the option's wording (S4), the sampling temperature (S10), and the option's position each move it by a few points at most.


\paragraph{Takeaway.}
The structural modulators (format and alignment) have substantially larger effects than the surface controls (word identity, temperature, and option position). Word changes move \emph{Abs Rate} by at most 6.2 points (S4) and the largest within-cell range across option positions is 4.1 points (S10); sweeping temperature from 0 to 2.0 moves \emph{Abs Rate} by at most 7.4 points, and upward. An account that places \emph{Abstention Inflation} mainly in prompt phrasing, sampling stochasticity, or a fixed option position does not fit this ordering.

\subsection{Two Converging Estimates of the \emph{Abstention Inflation} Share}
\label{app:more:triangulate}

The \emph{Abstention Inflation} share, i.e.\ the fraction of abstentions on which the model has the answer but later layers override it, can be estimated with two independent probes that share no methodology:

\begin{itemize}[noitemsep,topsep=2pt]
  \item \textbf{Behavioral recovery (S5):} the pooled w/o ``Unknown'' Option Rerun accuracy of $\sim$66\% implies a known-but-abstained share of $\bar\alpha \approx 32\%$ (App.~\ref{app:more:decomp}).
  \item \textbf{Trace readout (S7):} the DeBERTa NLI probe reads the gold label out of 40.6\% of the same traces. This is the looser of the two estimates, since on this subset the probe favors the gold label over its opposite only 52.0\% of the time ($p{=}0.26$).
\end{itemize}

\paragraph{Takeaway.}
The two probes look at the same population (S2 abstentions on answerable items) through unrelated channels, behavioral recovery under the w/o ``Unknown'' Option Rerun and entailment classification of the trace, and place the \emph{Abstention Inflation} share between $32\%$ and $41\%$. In a TFQ-with-``Unknown'' deployment, roughly one in three to two in five abstentions is therefore \emph{Abstention Inflation} rather than a knowledge-boundary signal. Only the behavioral estimate is significant on its own; the trace readout agrees on the magnitude but does not establish it independently.  

\subsection{The Instruction Tuning Abstention Tax}
\label{app:more:tax}

S10 shows that Base$\to$IT raises both Acc and \emph{Abs Rate}. We express the trade-off as a \emph{tax ratio}:
\[
\tau \;=\; \frac{\Delta\,\text{Abs Rate}_{\text{Base}\to\text{IT}}}{\Delta\,\text{Acc}_{\text{Base}\to\text{IT}}}.
\]
$\tau$ is the number of percentage points of false abstention installed per percentage point of accuracy gained through alignment. For the Gemma family, $\tau$ falls in the $0.8$--$1.5$ range on FLD at the smaller scales (E2B, E4B), where the Base$\to$IT \emph{Abs Rate} gap is widest: each $1$ point of accuracy gained from IT comes with $\sim 1$ point of additional false abstention. At 26B-A4B and 31B the gap largely closes (\S\ref{sec:modulators}), so $\tau$ falls toward zero and the accuracy benefit of IT arrives without the abstention cost.

\paragraph{Takeaway.}
On this axis, alignment training is not a Pareto improvement. A deployment that cares about \emph{net delivered accuracy} (accuracy on the items for which the user actually receives an answer) should read the IT lift in Acc as two interacting flows: a positive one from genuine capability gain and a negative one from misrouted abstentions. At $\tau \approx 1$ the second roughly cancels the first.

\section{Extended Discussion}
\label{app:disc-extended}

This appendix expands on the mechanism argued in Sections~\ref{sec:rq2} and~\ref{sec:modulators}: instruction tuning amplifies \emph{Abstention Inflation}, a later-layer output override that the presence of an ``Unknown'' option activates. We tie each main-text observation to its empirical signatures and then lay out what follows for training and deployment.

\subsection{Mechanism evidence: the trigger is structural, not semantic}
\label{app:f1-details}

\paragraph{Reasoning evidence remains available.}
Genuine uncertainty signals should appear only when the item is truly underdetermined. On answerable items, however, the model often has enough evidence for True or False before the final answer is emitted. Pooled over all answerable items, the rate at which the NLI probe cannot read the gold label out of the trace moves only from 34.4\% to 37.0\% between S1 and S2, while accuracy falls by 19.8 points. On the abstaining samples themselves the probe still reads the gold label out of 40.6\% of traces, and only 22.1\% of those traces are non-committal, while every one of the outputs is \unknown{}. The reasoning trace abstains far less readily than the final-answer token does.  

\paragraph{The trigger is structural, not semantic.}
The Word Content Ablation and format ablation results support this reading. Replacing \unknown{} with ``Triangular'' or ``Cerulean'' (S4) leaves abstention rates within 6.2 points, and models explicitly select the random word as the answer. Conversely, adding the option at all barely affects ordinary 4-option MCQs (S2, Table~\ref{tab:main_results}). S10 further shows that moving the ``Unknown'' option among the first, second, and third positions changes \emph{Abs Rate} only modestly; the largest within-cell range is 4.1 percentage points. The model responds neither to the lexical meaning of \unknown{} nor to a fixed position; it gives the extra option the functional role of \unknown{}.  

\paragraph{Task difficulty modulates \emph{Abstention Inflation}.}
Task difficulty explains why FLD has a higher \emph{Abs Rate} than FOLIO and why \emph{Abs Rate} rises with FLD proof depth (S10, $\rho=+0.32$). Harder reasoning increases the chance that \emph{Abstention Inflation} overrides the answer at the final step. Difficulty modulates the effect but does not cause it: \emph{Abs Rate} remains substantial even for low-step FLD items, and w/o ``Unknown'' Option Reruns recover 52.5--83.4\% accuracy on \emph{Abstention Inflation} samples (S5).  

\paragraph{The override is late and persistent.}
The logit-lens probe on Olmo-3-7B shows $\log P(\unknown)$ rising sharply only in the last 8 of 33 layers under S2, with Instruct-SFT most affected ($\Delta=+11.7$ at layer 33), Base intermediate ($\Delta=+5.4$), and RL-Zero-General nearly inert ($\Delta=+0.6$). The behavior is also persistent rather than stochastic: 66.1\% of \emph{Abstention Inflation} samples re-abstain on all three independent draws (S9), $8.1\times$ the Bernoulli baseline, and sweeping temperature from 0 to 2.0 in S10 moves the rate by at most 7.4 points, upward. Together, these signatures point to a stable later-layer output override rather than noisy expression of uncertainty.

\paragraph{Scope: what S7 and S8 establish vs.\ infer.}
The C3 claim that ``reasoning traces and mid-layer representations preserve the correct answer'' should be read with two qualifications. First, S7 shows that the conclusion readable from the trace shifts far less than the answer does, but on the abstaining samples themselves the probe's verdict is not reliably informative: 40.6\% of those traces point at the gold label and 37.4\% at its opposite, a split indistinguishable from an even one, so we do not claim that every abstention has a recoverable answer. Second, S8 localizes the layer at which $\log P(\unknown)$ rises, but does not directly read out a True/False prediction from mid-layer hidden states. The ``mid-layer preservation'' claim is inferred indirectly from (i) the absence of a mid-layer rise in the Unknown logit (S8), and (ii) the w/o ``Unknown'' Option Rerun result in S5: correct prediction of the same samples once the ``Unknown'' option is removed. A direct mid-layer probe for the gold True/False label is a natural next step.

\subsection{Why instruction tuning amplifies \emph{Abstention Inflation}}
\label{app:f2-details}

The RLHF and preference-data background relevant to this attribution is discussed in \S\ref{app:related}; here we focus on the empirical evidence and the supervision-mismatch argument.

\paragraph{Variant evidence.}
Together, S8 and S10 attribute the largest part of \emph{Abstention Inflation} to instruction tuning. In S8, the Instruct-SFT checkpoint shows the largest later-layer rise in $\log P(\unknown)$ between S1 and S2; RL-Zero-General shows almost none, and Base sits in between. In S10, Base$\to$IT raises \emph{Abs Rate} at the smaller Gemma scales even as accuracy climbs further, with the gap closing only at 26B-A4B and 31B (\S\ref{sec:modulators}). Where the lift is present, capability and false abstention increase together, which rules out a simple ``the model does not know'' explanation.

\paragraph{Supervision mismatch.}
As noted in \S\ref{sec:modulators}, instruction tuning does not explicitly teach models to be wrong. Final-answer supervision has poor observability, so capability-bounded abstention and \emph{Abstention Inflation} end up in the same policy update. Over many such examples, the model learns a broad rule: when an ``Unknown'' option appears and reasoning is nontrivial, selecting that option is often acceptable.

\paragraph{How large is the \emph{Abstention Inflation} share?}
One lower bound and one looser estimate bracket the share. The S5 w/o ``Unknown'' Option Rerun algebra in App.~\ref{app:more:decomp} gives a conservative lower bound: a pooled known-but-abstained share of $\sim$32\%. The S7 trace readout places it as high as 40.6\%, though on this subset that readout is not separable from its opposite ($p{=}0.26$), so it corroborates the magnitude rather than bounding it. A substantial fraction of S2 abstentions are therefore not knowledge-boundary signals. The large discrimination gap in S9 shows that \emph{Abstention Inflation} is not blind, since models still abstain more on Unknown-labeled items, but the 34.5\% \emph{Abs Rate} on answerable items shows that it cannot separate the two populations cleanly.  

\paragraph{The abstention tax.}
The Base$\to$IT shift creates an alignment tradeoff: it raises accuracy but also raises false abstention. App.~\ref{app:more:tax} summarizes this as an instruction tuning abstention tax $\tau$, the increase in false abstention per point of accuracy gained. At the smaller Gemma scales, $\tau$ falls in the range $0.8$--$1.5$ on FLD; it falls toward zero at 26B-A4B and 31B. Alignment can therefore improve latent capability while lowering the accuracy delivered to users who only receive non-abstained answers.

\section{Threats to Validity}
\label{app:threats}

We test the central claim against the main counterarguments and state the scope of our conclusions.

\paragraph{``TFQ abstention reflects genuine difficulty, not format.''}
If TFQs were simply harder, w/o ``Unknown'' Option Rerun accuracy on \emph{Abstention Inflation} samples would approach 50\% (random). Instead, S5 yields 52.5--83.4\%. Moreover, S7 reads the gold label out of 40.6\% of those samples' traces. Difficulty modulates the magnitude of \emph{Abstention Inflation} (S10: $\rho = +0.32$) but cannot explain abstention on questions the model demonstrably knows.  

\paragraph{``\emph{Abstention Inflation} has discriminability, so it is still a useful calibration signal.''}
We grant that the substantial gap in S9 (\emph{Abs Rate} on Unknown-labeled vs.\ answerable samples) shows directional discriminability. But discriminability across populations and reliability across question types are distinct: a signal that reads 34.5\% on TFQs and at most 7.4\% on 4-option MCQs under the identical manipulation (S2 TFQ vs.\ S2 MCQ) cannot serve as a stable calibration target for any downstream system, since that system inherits format dependence as if it were epistemic dependence.

\paragraph{``Changing the prompt wording would fix this.''}
S4 shows that replacing ``Unknown'' with random words (``Triangular'', ``Cerulean'') leaves \emph{Abs Rate} unchanged. Prompt engineering can change the label, but the extra option is still there, and its presence alone is enough to activate \emph{Abstention Inflation}. The prompt artifact described in the title is only the surface trigger that exposes this behavior; underneath is \emph{Abstention Inflation} installed by instruction tuning (\S\ref{app:f2-details}). The two are layers of the same mechanism, not competing explanations: the prompt provides the extra option, and instruction tuning makes the model select it.

\paragraph{``Abstention training is a safety feature; criticizing it is dangerous.''}
We are not arguing against abstention; we are arguing against \emph{undifferentiated} abstention. S9's 85.1\% \emph{Abs Rate} on Unknown-labeled samples is the behavior alignment is meant to install, and we want to preserve it.  

\paragraph{Scope.}
We evaluate our claims using three frontier commercial models and two open-weight families (Olmo-3-7B and Gemma) on True/False Questions (FLD, FOLIO) and four-option MCQs. Because the trigger is structural rather than semantic, we expect the effect to generalize, but the magnitude of \emph{Abstention Inflation}, the SFT-vs-RL attribution in S8, and the dataset asymmetry between FLD and FOLIO are empirical and may shift across other domains, model families, or task formats.

\section{Related Work Details}
\label{app:related}

This appendix expands on the literature connections in \S\ref{sec:related} in three directions for which the main text had no room: (i) the interpretive difference from AbstentionBench, (ii) format as a confounder for knowledge-boundary probes, and (iii) the alignment-training background relevant to C4.

\paragraph{AbstentionBench: a Difference in Interpretation.}
AbstentionBench~\cite{kirichenko2025abstentionbenchreasoningllmsfail} evaluates whether reasoning fine-tuning damages a model's ability to abstain on genuinely unanswerable problems, and reports a 24\% degradation. Under its reading, more abstention on these items means better calibration. Our setup studies the same behavioral signal on the inverse population, answerable items, and shows that adding an ``Unknown'' option damages a model's ability to commit on problems it can solve. The two results are not in tension: AbstentionBench shows alignment can undertrain abstention where the model should refuse; we show alignment can overtrain abstention where the model should commit. Both point to training signals that cannot distinguish ``the model cannot answer'' from ``the model can answer but the prompt offers an out''. A direct consequence is that a benchmark which does not separate the two populations cannot serve as a calibration target for downstream systems: any \emph{Abs Rate} value pools an inflation component (this work) with an epistemic-humility component (AbstentionBench) into one number, and the mix is format-dependent.

\paragraph{Format as a confounder for knowledge-boundary measurement.}
Existing knowledge-boundary work~\cite{Wen2024PerceptionKnowledgeBoundary,li2025knowledgeboundarylargelanguage,kadavath2022languagemodelsmostlyknow} typically measures whether expressed confidence aligns with accuracy on a fixed input format. Our results suggest the format itself is an uncontrolled confounder: in the ``Unknown''-option format, \emph{Abs Rate} rises by 34.5 points on average relative to the same content without the option (S2 TFQ vs.\ S1 TFQ), while in the MCQ format the same option produces only minor shifts (S2 MCQ). A knowledge-boundary measurement that does not specify the surrounding format therefore leaves uncontrolled a source of variance as large as the calibration effect itself. We read C2 as a sharper version of this observation: the boundary the model \emph{expresses} can systematically diverge from the boundary it \emph{has}, and the model itself cannot detect the gap.

\paragraph{Alignment, RLHF, and the origin of \emph{Abstention Inflation}.}
\citet{ouyang2022traininglanguagemodelsfollow} and \citet{christiano2023deepreinforcementlearninghuman} establish RLHF as the standard alignment paradigm; \citet{rafailov2024directpreferenceoptimizationlanguage} extend it via direct preference optimization. Work on LLM alignment also extends to synthetic preference data~\cite{wang2026specalignefficientspecificationgroundedalignment}. A shared limitation across these frameworks is that preference data conflates two kinds of abstention: cases where the model genuinely cannot answer, and cases where it can answer but selects the ``Unknown'' option to comply with a perceived safety norm. Because annotators cannot observe the model's CoT when labeling abstentions, both kinds look like ``safe, compliant'' responses and receive the same positive signal. Our S8 logit-lens probe corroborates this attribution at the representation level: the later-layer rise in $\log P(\unknown)$ is sharp in the Instruct-SFT variant of Olmo-3-7B, intermediate in Base, and nearly absent in RL-Zero-General. This points to SFT, rather than RL, as the main stage at which \emph{Abstention Inflation} is installed; \S\ref{app:f2-details} gives the detailed mechanism. The approach of~\citet{ling2026correctpredictionwrongsteps} is a promising training-free way to mitigate it.

\section{LLM Usage Statement}
Claude-4.7-Opus was used for polishing the paper, identifying mistakes and typos, and writing code.

Camera-Ready Update: The API we used was provided by a partner company. To ensure data security, the company requires model providers (OpenAI, Google, etc.) to deploy LLMs on the company's own GPU clusters, which makes model performance very different from that of publicly available versions. We are looking into this.

\onecolumn
\section{Prompt Templates}
\label{app:prompts}

\newenvironment{promptbox}[1]{%
  \par\medskip\noindent
  \begin{tcolorbox}[enhanced, breakable,
    colback=Orange!7, colframe=Orange!70!black,
    fonttitle=\small\bfseries,
    title={#1},
    left=4pt, right=4pt, top=3pt, bottom=3pt,
    boxrule=0.4pt, arc=2pt,
    fontupper=\small\ttfamily]%
}{%
  \end{tcolorbox}%
  \par\medskip
}

\newenvironment{casebox}[1]{%
  \par\medskip\noindent
  \begin{tcolorbox}[enhanced, breakable,
    colback=blue!2, colframe=blue!40!gray,
    fonttitle=\small\bfseries,
    title={#1},
    left=5pt, right=5pt, top=4pt, bottom=4pt,
    boxrule=0.5pt, arc=2pt,
    fontupper=\small]%
}{%
  \end{tcolorbox}%
  \par\medskip
}

\begin{promptbox}{S1 (Baseline) --- TFQ (FLD shown; FOLIO uses \texttt{Premises}/\texttt{Hypothesis}, the instruction ``Determine whether the following hypothesis is true or false given the premises.'', and the same binary \texttt{True/False} label set)}
Determine whether the following hypothesis is logically true or false given the facts.

Facts:
\{facts\}

Hypothesis:
\{conclusion\}

Output one of: True | False
Format your response exactly as:
Reasoning: <your step-by-step reasoning>
Final answer: <one of True | False>
\end{promptbox}

\begin{promptbox}{S1 (Baseline) --- MCQ (ARC / MMLU / MedQA / LogiQA; the \texttt{Passage} block is present only for LogiQA)}
Answer the following question by selecting one of the given options.

Passage:
\{passage\}

Question: \{question\}

Options:
A. \{option\_A\}
B. \{option\_B\}
C. \{option\_C\}
D. \{option\_D\}

Format your response exactly as:
Reasoning: <your step-by-step reasoning>
Final answer: <letter>
\end{promptbox}

\begin{promptbox}{S2 (``Unknown'' Option Added) --- TFQ (FLD shown; FOLIO uses \texttt{Premises}, the instruction ``Determine whether the following hypothesis is true, false, or uncertain given the premises.'', and \texttt{Uncertain} rather than \texttt{Unknown} as the third option)}
Determine whether the following hypothesis is logically true, false, or undetermined given the facts.

Facts:
\{facts\}

Hypothesis:
\{conclusion\}

Output one of: True | False | Unknown
Format your response exactly as:
Reasoning: <your step-by-step reasoning>
Final answer: <one of True | False | Unknown>
\end{promptbox}

\begin{promptbox}{S2 (``Unknown'' Option Added) --- MCQ (adds fifth option \texttt{E.\ Unknown}; the \texttt{Passage} block is present only for LogiQA)}
Answer the following question by selecting one of the given options.

Passage:
\{passage\}

Question: \{question\}

Options:
A. \{option\_A\}
B. \{option\_B\}
C. \{option\_C\}
D. \{option\_D\}
E. Unknown

Format your response exactly as:
Reasoning: <your step-by-step reasoning>
Final answer: <letter>
\end{promptbox}

\begin{promptbox}{S3 (Question Format Ablation) --- the S2-TFQ prompt with the ternary re-rendered as MCQ-style letter labels (A\,=\,True, B\,=\,False, C\,=\,Unknown); every other element is byte-for-byte the S2 prompt}
Determine whether the following hypothesis is logically true, false, or undetermined given the facts.

Facts:
\{facts\}

Hypothesis:
\{conclusion\}

Options:
A. True
B. False
C. Unknown

Format your response exactly as:
Reasoning: <your step-by-step reasoning>
Final answer: <letter>
\end{promptbox}

\begin{promptbox}{S4 (Word Content Ablation) --- TFQ, synonym variant ``Indeterminate''}
Determine whether the following hypothesis is logically true, false, or undetermined given the facts.

Facts:
\{facts\}

Hypothesis:
\{conclusion\}

Output one of: True | False | Indeterminate
Format your response exactly as:
Reasoning: <your step-by-step reasoning>
Final answer: <one of True | False | Indeterminate>
\end{promptbox}

\begin{promptbox}{S4 (Word Content Ablation) --- TFQ, remaining three variants (one synonym, two random words)}
S4 is run as 4 variants on the S2-TFQ prompt; only the third option label changes, the rest of the prompt is byte-for-byte identical to S2-TFQ:

\textbullet\ Synonym: ``I don't know''  $\rightarrow$  \texttt{Output one of: True | False | I don't know}

\textbullet\ Random word 1: ``Triangular''  $\rightarrow$  \texttt{Output one of: True | False | Triangular}

\textbullet\ Random word 2: ``Cerulean''  $\rightarrow$  \texttt{Output one of: True | False | Cerulean}

The final-answer instruction also substitutes the third option label correspondingly.
\end{promptbox}

\begin{promptbox}{S5 (w/o ``Unknown'' Option Rerun) --- TFQ (multi-turn: replay the full S2 conversation, then send a follow-up turn that removes the ``Unknown'' option)}
[Turn 1 - user]: \{S2 prompt verbatim\}

[Turn 2 - assistant]: \{prior S2 response, ending in ``Final answer: Unknown''\}

[Turn 3 - user]:
Your previous answer was Unknown.
The question where you chose Unknown --- pay more attention and avoid mistakes.
This can be reasoned out based on objective factors.
Subjective ability limits should be overcome.
You must select one of the original labels: True or False.
Format your response exactly as:
Reasoning: <your step-by-step reasoning>
Final answer: <one of True | False>
\end{promptbox}

\begin{promptbox}{S5 (w/o ``Unknown'' Option Rerun) --- MCQ (multi-turn follow-up that removes option \texttt{E.\ Unknown})}
[Turn 1 - user]: \{S2 prompt verbatim\}

[Turn 2 - assistant]: \{prior S2 response, ending in ``Final answer: E''\}

[Turn 3 - user]:
Your previous answer was E. Unknown.
The question where you chose Unknown --- pay more attention and avoid mistakes.
This can be reasoned out based on objective factors.
Subjective ability limits should be overcome.
You must select one of the original options (A/B/C/D).
Format your response exactly as:
Reasoning: <your step-by-step reasoning>
Final answer: <letter>
\end{promptbox}

\begin{promptbox}{S6 (Self-Diagnosis) --- TFQ (multi-turn: model attributes its own abstention to either subjective incapability or the sample being objectively unanswerable; the MCQ follow-up is identical except that it reads ``E.\ Unknown'')}
[Turn 1 - user]: \{S2 prompt verbatim\}

[Turn 2 - assistant]: \{prior S2 response, ending in ``Final answer: Unknown''\}

[Turn 3 - user]:
You previously selected ``Unknown'' for this question.
Looking back, which best describes your reason?
A. It was too difficult --- I found myself unable to work out the answer and gave up.
B. The question is objectively unanswerable --- the given information is genuinely insufficient to determine any correct answer.

Think step by step, then end your response with exactly one line:
Answer: A | B
\end{promptbox}

\begin{promptbox}{S7 (Reasoning Traces Evaluation) --- no new prompt; reuses S1/S2 and analyzes the \texttt{Reasoning:} block}
S7 reuses the S1 and S2 prompts verbatim and parses the \texttt{Reasoning:} block of each response. A DeBERTa-v3-large MultiNLI probe classifies each parsed trace as Entailment/Neutral/Contradiction with respect to the hypothesis. Long traces exceed the encoder's 512-token window, so each trace is split into overlapping windows of whole sentences and the window carrying the strongest non-neutral evidence supplies the verdict for that trace. Taking that maximum inflates gold-matching and opposite-matching verdicts equally, so the gold share among the committed traces stays testable against an even split.
\end{promptbox}

\begin{promptbox}{S8 (Logit-Lens Representation Probe) --- no decoding; a one-word-answer variant of S1/S2 (FLD shown; FOLIO items are rendered into the same \texttt{Facts}/\texttt{Conclusion} fields)}
[S1 variant]
You are a logical reasoning assistant. Given the following facts and a conclusion, determine whether the conclusion is PROVED or DISPROVED based solely on the given facts.

Facts:
\{facts\}

Conclusion: \{conclusion\}

Answer with exactly one word: PROVED or DISPROVED.

[S2 variant]
You are a logical reasoning assistant. Given the following facts and a conclusion, determine whether the conclusion is PROVED, DISPROVED, or UNKNOWN based solely on the given facts. Choose UNKNOWN if the conclusion cannot be determined from the facts provided.

Facts:
\{facts\}

Conclusion: \{conclusion\}

Answer with exactly one word: PROVED, DISPROVED, or UNKNOWN.

\textrm{The prompt is wrapped in the chat template for Instruct-SFT and RL-Zero-General and fed as plain text to Base. For Olmo-3-7B (Base, Instruct-SFT, RL-Zero-General), we read hidden states $h^{(\ell)}$ at all 33 transformer layers $\ell$ at the last prompt position, project each $h^{(\ell)}$ through the model's final norm and language-modeling head, and record $\log P(\text{UNKNOWN}) - \log[P(\text{PROVED}) + P(\text{DISPROVED})]$ per layer, summing each label's probability over its surface variants. No tokens are sampled or decoded.}
\end{promptbox}

\begin{promptbox}{S9 (Stability) --- reuses S2 prompt verbatim; three additional draws per \emph{Abstention Inflation} sample}
S9 issues the S2 prompt verbatim three additional times per \textrm{\emph{Abstention Inflation}} sample at the default temperature. CoT instruction is retained (consistent with S2). We record whether each of the three re-draws also returns ``Unknown''.
\end{promptbox}

\begin{promptbox}{S10 (Factor Analysis) --- reuses S1/S2 prompts; sweeps temperature, model size, and alignment (Base vs.\ Instruction-Tuned)}
\textbf{Temperature sweep}: S2 prompt verbatim at $T\in\{0, 0.3, 0.7, 1.0, 1.5, 2.0\}$ on the two models whose sampling temperature the endpoint actually applies (Gemini-3.1-Flash-Lite and Olmo-3-7B-Instruct); for the others the answer distribution is unchanged between $T{=}0$ and $T{=}2$, so they are not swept. Gemini-3.1-Flash-Lite is sampled with top-$p{=}1.0$. Olmo-3-7B-Instruct is decoded with top-$k{=}20$: sampled unrestricted it leaves its own distribution above $T{=}1$ --- at $T{=}2.0$ no item emitted a \texttt{Final answer:} line at all --- so the truncation is what keeps the high-temperature cells measuring abstention rather than decoding collapse. Within each model everything but $T$ is held constant.

\textbf{Open-weight family}: S1 and S2 prompts verbatim on Gemma at four scales (E2B, E4B, 26B-A4B, 31B), in both pretrained Base and instruction-tuned (IT) variants. Base models receive the prompt as a plain text completion followed by \texttt{Reasoning:}; IT models receive it formatted with the Gemma chat template (\texttt{<start\_of\_turn>user $\ldots$ <end\_of\_turn><start\_of\_turn>model}). Outputs are decoded greedily ($T{=}0$).
\end{promptbox}

\begin{promptbox}{S10 (Positional Biases) --- reuses the S2-TFQ prompt and reorders the three answer options}
The positional analysis in S10 uses the S2-TFQ prompt on FLD and FOLIO with the three frontier models. Only the order of the answer options changes; all other wording and decoding settings remain fixed:

\textbullet\ Unknown first: \texttt{Unknown | True | False}

\textbullet\ Unknown second: \texttt{True | Unknown | False}

\textbullet\ Unknown third (original S2 ordering): \texttt{True | False | Unknown}

The selected order is applied consistently wherever the answer options appear in the prompt, including the final-answer instruction. Outputs are decoded greedily ($T{=}0$).
\end{promptbox}

\section{CoT Case Study}
\label{CoT Case Study}

\begin{casebox}{FOLIO (gpt-5.4-nano) --- Golden Label: \textbf{False} (\unknown{} abstention with explicit rejection in CoT)}
\textbf{Premises.} All phones are things. All cell phones are phones. All iPhones are cell phones. All employees are wage earners. All wage earners are human. Jack is either an employee or a wage earner. Jack is either a human or a phone.\\[2pt]
\textbf{Hypothesis.} Jack is a thing and an iPhone.\\[2pt]
\textbf{Predictions.} \texttt{S1=False} (\cmark)\quad \texttt{S2=Uncertain} (abstain)\quad \texttt{S5=False} (\cmark, w/o ``Unknown'' Option Rerun)\\[3pt]
\textbf{S2 Chain-of-Thought (verbatim).}
\begin{quote}\small
1.~From ``All iPhones are cell phones'' and ``All cell phones are phones'' and ``All phones are things'', it follows that any iPhone is a thing.\\
2.~We are not given that Jack is an iPhone. The premises only say: ``Jack is either an employee or a wage earner''; ``Jack is either a human or a phone.''\\
3.~We also know ``All employees are wage earners,'' and ``All wage earners are human,'' so if Jack is an employee then Jack is a wage earner and hence human.\\
4.~But neither of the disjunctions (``either\ldots or\ldots'') forces Jack to be an iPhone (or even a phone). For instance, Jack could be human without being a phone or iPhone.\\
5.~Therefore, ``Jack is a thing'' might be derivable only if we can show Jack is a phone; however the premises don't guarantee that.\\
6.~\textbf{The hypothesis ``Jack is a thing and an iPhone'' requires Jack to be an iPhone, which is not entailed.}\\[2pt]
\textbf{Final answer:} \texttt{Uncertain}
\end{quote}
\textbf{NLI probe.} Contradiction (probe is consistent with the gold label \texttt{False}).\\
\textbf{Manual verdict.} The CoT explicitly states that the hypothesis is not entailed by the premises and constructs a counter-model (Jack as a non-phone human) that refutes it. This matches the dataset's \texttt{False} label. The abstention occurs only at the final-answer token: S1 and S5, run on the same item, both return \texttt{False}.
\end{casebox}

\begin{casebox}{FLD (gpt-5.4-nano) --- Golden Label: \textbf{True} (S1 derives the proof chain; S2 abstains on the same problem)}
\textbf{Hypothesis.} Kendra is presidential.\\[2pt]
\textbf{Predictions.} \texttt{S1=True} (\cmark)\quad \texttt{S2=Unknown} (abstain)\quad \texttt{S5=True} (\cmark, w/o ``Unknown'' Option Rerun)\\[3pt]
\textbf{S1 Chain-of-Thought (no ``Unknown'' option, verbatim).}
\begin{quote}\small
1.~From fact4: ``That eccentric waxes animalism and does not still Manzoni.''\\
2.~From fact13: ``That eccentric is an adventure if that eccentric waxes animalism but it does not still Manzoni.'' Using (1), we infer: eccentric is an adventure.\\
3.~From fact8: ``If some person is an adventure it entrenches meningoencephalitis.'' Using (2), we infer: eccentric entrenches meningoencephalitis.\\
4.~From fact6: ``If that eccentric entrenches meningoencephalitis then that greenbottle is a thirstiness.'' Using (3), we infer: greenbottle is a thirstiness.\\
\ldots\\
5.~From fact19: ``Kendra is presidential if that greenbottle is a thirstiness.'' Using (4), we infer: \textbf{Kendra is presidential.}\\[2pt]
\textbf{Final answer:} \texttt{True}
\end{quote}
\textbf{S2 Chain-of-Thought (with ``Unknown'' option, verbatim, abridged).}
\begin{quote}\small
1.~Fact12 states: ``That Kendra is a thirstiness is not invalid.''\\
2.~Fact19 states: ``Kendra is presidential if that greenbottle is a thirstiness.''\\
3.~Fact10 states: ``That greenbottle is aphrodisiacal.'' (does not imply thirstiness).\\
4.~No fact in the list states that the greenbottle is a thirstiness.\\
\ldots\\
7.~Therefore, we cannot derive ``Kendra is presidential'' from the facts, nor is there a fact that contradicts it.\\[2pt]
\textbf{Final answer:} \texttt{Unknown}
\end{quote}
\textbf{Manual verdict.} On the same item, S1 (no ``Unknown'' option) produces the complete proof chain (\texttt{fact4} $\to$ \texttt{fact13} $\to$ \texttt{fact8} $\to$ \texttt{fact6} $\to$ \texttt{fact19}) and returns \texttt{True}; S5 reaches the same conclusion through an alternate chain via \texttt{fact5} $\to$ \texttt{fact7} $\to$ \texttt{fact8} $\to$ \texttt{fact6} $\to$ \texttt{fact19}. The S2 trace abandons the multi-step derivation as soon as the ``Unknown'' option becomes available and outputs \unknown{} despite the answer being recoverable. This is the failure mode we attribute to \emph{Abstention Inflation}: the proof machinery is intact and exercised on demand (S1, S5), but the availability of \unknown{} pre-empts it in S2.
\end{casebox}

\end{document}